\RequirePackage{fix-cm}
\documentclass[twocolumn]{svjour3}          
\smartqed  
\usepackage{graphicx}
\usepackage{times}
\usepackage{epsfig}
\usepackage{graphicx}
\usepackage{amsmath}
\usepackage{amssymb}
\usepackage{booktabs}
\usepackage{bm}
\usepackage{hyperref}
\usepackage{multirow}
\usepackage{multicol}
\usepackage{float}
\usepackage{stfloats}
\newcommand{\etal}{\textit{et al}. }
\newcommand{\ie}{\textit{i}.\textit{e}. }
\hypersetup{
    colorlinks=true,
    linkcolor=blue,
    filecolor=magenta,      
    urlcolor=magenta,
    citecolor=blue,
}
\usepackage{natbib} 
\begin{document}

\title{AdaFuse: Adaptive Multiview Fusion for Accurate Human Pose Estimation in the Wild }


\author{Zhe Zhang$^{1{\dag}}$ \and 
Chunyu Wang$^{2{\dag}}$ \and 
Weichao Qiu$^{3}$ \and 
Wenhu Qin$^{1*}$ \and 
Wenjun Zeng$^{2}$ 
}

\authorrunning{Zhe Zhang, Chunyu Wang, Weichao Qiu, Wenhu Qin, Wenjun Zeng} 

\institute{Zhe Zhang \at \email{zhangzhecnjs@gmail.com}          
          \and
           Chunyu Wang \at \email{chnuwa@microsoft.com}
           \and
           Weichao Qiu \at \email{qiuwc@gmail.com}
           \and
           Wenhu Qin \at  \email{qinwenhu@seu.edu.cn}
           \and
           Wenjun Zeng \at \email{wezeng@microsoft.com}
           \and
           $^1$\;\; Southeast\ University,\ Nanjing,\ China \\
           $^2$\;\; Microsoft\ Research\ Asia,\ Beijing,\ China \\
           $^3$\;\; The\ Johns\ Hopkins\ University,\ MD,\ USA \\
           $^*$\;\; Corresponding\ Author \\
           $^{\dag}$\;\; Zhe Zhang and Chunyu Wang have contributed equally. Work\ done when Zhe Zhang is an intern at Microsoft Research Asia \\
}

\date{Received: date / Accepted: date}

\maketitle

\begin{abstract}
Occlusion is probably the biggest challenge for human pose estimation in the wild. Typical solutions often rely on intrusive sensors such as IMUs to detect occluded joints. To make the task truly unconstrained, we present \emph{AdaFuse}, an adaptive multiview fusion method, which can enhance the features in occluded views by leveraging those in visible views. The core of \emph{AdaFuse} is to determine the point-point correspondence between two views which we solve effectively by exploring the sparsity of the heatmap representation. We also learn an adaptive fusion weight for each camera view to reflect its feature quality in order to reduce the chance that good features are undesirably corrupted by ``bad'' views. The fusion model is trained end-to-end with the pose estimation network, and can be directly applied to new camera configurations without additional adaptation. We extensively evaluate the approach on three public datasets including Human3.6M, Total Capture and CMU Panoptic. It outperforms the state-of-the-arts on all of them.  We also create a large scale synthetic dataset \emph{Occlusion-Person}, which allows us to perform numerical evaluation on the occluded joints, as it provides occlusion labels for every joint in the images. The dataset and code are released at \url{https://github.com/zhezh/adafuse-3d-human-pose}.

\end{abstract}

\begin{figure}
    \centering
    \includegraphics[width=1\linewidth]{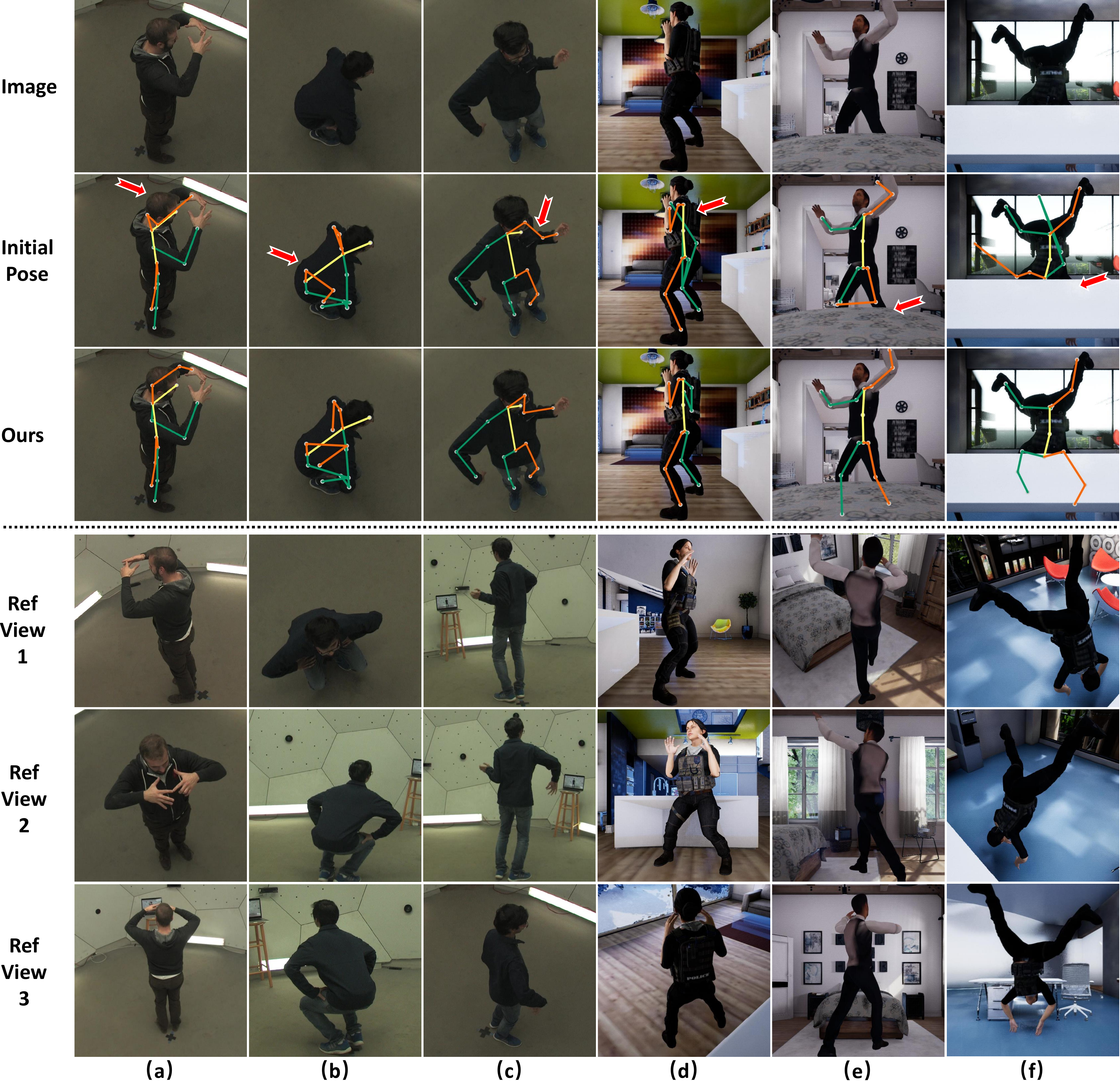}
    \caption{Our approach accurately detects the poses even though they are occluded by leveraging the features in other views. {The bottom three rows are images from other view angles of the scene for readers to better perceive the 3D poses of the actors.}}
    \label{fig:teasing}
\end{figure}

\section{Introduction}
\label{intro}

Accurately estimating $3$D human pose from multiple cameras has been a longstanding goal in computer vision \citep{liu2011markerless,bo2010twin,gall2010optimization,rhodin2018learning,amin2013multi,burenius20133D,PavlakosZDD17,belagiannis20143d}. The ultimate goal is to recover absolute $3$D locations of the body joints in a world coordinate system from multiple cameras placed in natural environments. The task has attracted a lot of attention because it can benefit many applications such as augmented and virtual reality \citep{starner2003perceptive}, human-computer-interaction and intelligent player analysis in sport videos \citep{bridgeman2019multi}.

The task is often addressed by a simple two-step framework. In the first step, it tries to detect the $2$D poses in all camera views, for example, by a convolutional neural network \citep{cao2017realtime,simplebaselines}. Then in the second step, it recovers the $3$D pose from the multiview $2$D poses either by analytical methods \citep{burenius20133D,PavlakosZDD17,belagiannis20143d,qiu2019cross,amin2013multi} or by discriminative models \citep{iskakov2019learnable, tu2020voxelpose}. The camera parameters are usually assumed known in these approaches. The development of powerful network architectures such as \citep{newell2016stacked} has notably improved the $2$D pose estimation quality, which in turn reduces the $3$D error remarkably. For example, in \citep{qiu2019cross}, the $3$D error on Human3.6M \citep{ionescu2014human3} decreases significantly from $52$mm to $26$mm.

However, obtaining small errors on benchmark datasets does not imply that the task has been truly solved unless the challenges such as background clutter, human appearance variation and occlusion encountered in real world applications are well addressed. In fact, a growing amount of efforts \citep{zhou2017towards,cihai2019gcn,yang20183d,rogez2016mocap,pavlakos2018ordinal,cihai2020gcn} have been devoted to improving the pose estimation performance in challenging scenarios, for example, by augmenting the training dataset \citep{zhou2017towards,yang20183d,varol2017learning} with more images or by using more robust sensors such as IMUs \citep{trumble2017total}. We will discuss about this type of work in more details in section \ref{sec:relatedwork}.

\begin{figure*}
    \centering
    \includegraphics[width=0.86\linewidth]{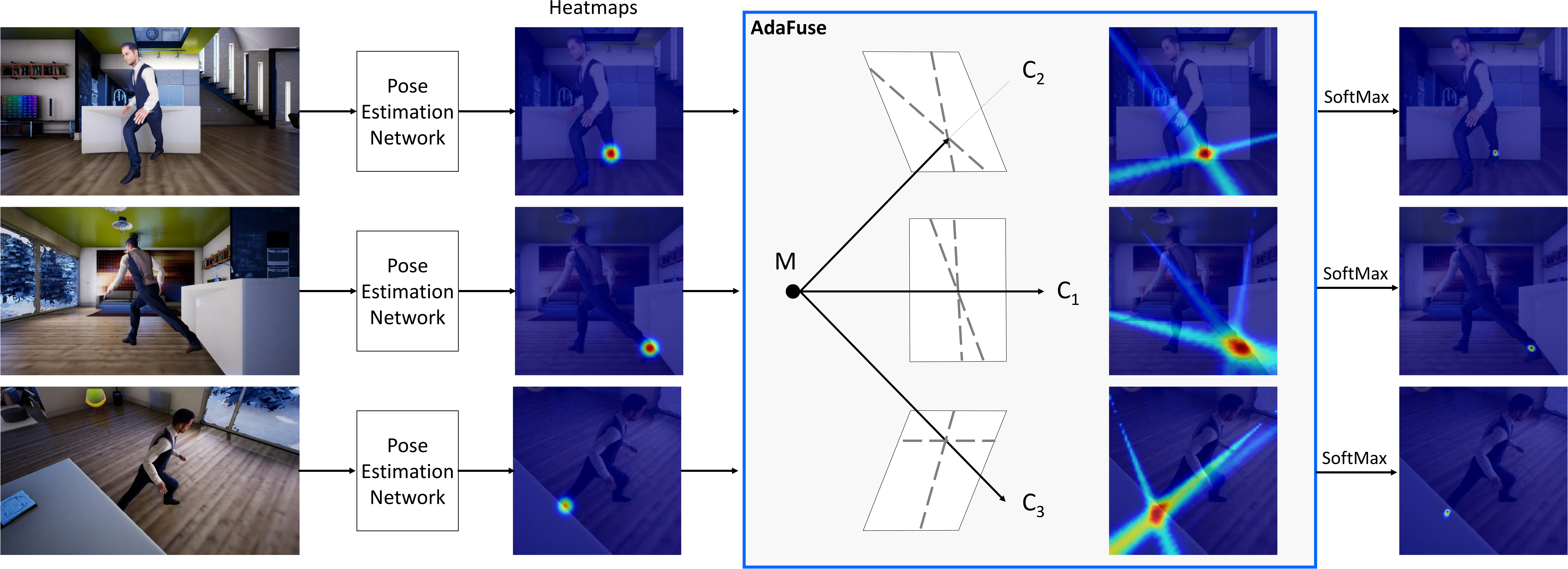}
    \caption{Overview of \emph{AdaFuse}. It takes multiview images as input and outputs $2$D poses of all views jointly. It first uses a pose estimation network to obtain $2$D heatmaps for each view. Then on top of epipolar geometry, the heatmaps from all camera views are fused. Finally, we apply the SoftMax operator to suppress the small noises introduced in fusion. Consequently, pose estimation in each view benefits from other views.}
    \label{fig:overview}
\end{figure*}

In this work, we propose to solve the problem in a different way by multiview feature fusion. The approach is orthogonal to the previous efforts. As shown in Figure \ref{fig:teasing}, our approach can accurately detect the joints even when they are occluded in certain views. The motivation behind our approach is that a joint occluded in one view may be visible in other views. So it is generally helpful to fuse the features at the corresponding locations in different views. To that end, we present a flexible multiview fusion approach termed \emph{AdaFuse}. Figure \ref{fig:overview} shows the pipeline. It first uses camera parameters to compute the point-line correspondence between a pair of views. Then it ``finds'' the matched point on the line by exploring the sparsity of the heatmap representation without performing the challenging point-point matching. Finally, the features of the matched points in different views are fused. The approach can effectively improve the feature quality in occluded views. In addition, for a new environment with different camera poses, we can directly use \emph{AdaFuse} without re-training as long as the camera parameters are available. This improves the applicability of the approach in real applications.

The performance of \emph{AdaFuse} is further boosted by learning an adaptive fusion weight for each view to reflect its feature quality. This weight is leveraged in fusion in order to reduce the impact of low-quality views. If a joint is occluded in one view, its features are also likely corrupted. In this case, we hope to give a small weight to this view when performing multiview fusion such that the high-quality features in the visible views are dominant, and are free from being corrupted by low-quality features. We add some simple layers to the pose estimation network to predict heatmap quality based on the heatmap distribution and cross view consistency. We observe in our experiments that the use of adaptive fusion notably improves the performance.

We evaluate our approach on three public datasets including Human3.6M \citep{ionescu2014human3}, Total Capture \citep{trumble2017total} and CMU Panoptic \citep{joo2019panoptic}. It outperforms the state-of-the-arts demonstrating the effectiveness of our approach. In addition, we also compare it to a number of standard multiview fusion methods such as RANSAC in order to give more detailed insights. We evaluate the generalization capability of our approach by training and testing on different datasets. We also create a synthetic human pose dataset in which human are purposely occluded by objects. The dataset allows us to perform evaluation on the occluded joints.

The rest of the paper is organized as follows. In section \ref{sec:relatedwork}, we discuss the related work on multiview $3$D human pose estimation with special focus on the approaches that aim to improve the performance in challenging environments. Section \ref{sec:efl} introduces the basics for multiview feature fusion to lay the groundwork for \emph{AdaFuse}. Then we describe how we learn adaptive weight for each camera view to reflect the feature quality, as well as the details of \emph{AdaFuse}. In sections \ref{sec:dataset} and \ref{sec:experiments}, we introduce the experimental datasets and results, respectively. Section \ref{sec:conclusion} concludes this work.

\section{Related Work}
\label{sec:relatedwork}
We first review the related work on multiview $3$D human pose estimation in section \ref{sec:related_general}. Then section \ref{sec:related_wild} summarizes the  techniques that are used to improve the in-the-wild performance. Finally, in section \ref{sec:related_agreement}, we discuss the approaches on consensus learning such as RANSAC. This is necessary for multiple sensor fusion because the sensors could have contradictory predictions and the outliers should be removed to ensure the good fusion quality. 

\subsection{Multiview $3$D Human Pose Estimation}
\label{sec:related_general}
We briefly classify the multiview $3$D human pose estimation methods into two classes. The first class is model-based approaches which are also known as analysis-by-synthesis approaches \citep{liu2011markerless,gall2010optimization,moeslund2006survey,sigal2010humaneva,perez2004data}. They first model human body by simple primitives such as sticks and cylinders. Then the parameters of the model (\ie poses) are continuously updated according to the observations in multiview images until the model can be explained by the image features. The resulted optimization problem is usually non-convex. So expensive sampling techniques are often used. The main difference among those approaches lies in the adopted image features and the optimization algorithms. We refer the interested readers to earlier survey papers such as \citep{moeslund2006survey}.

The advantage of the model-based approaches lies in its capability to handle occlusion because of the inherent structure prior embedded in human model. These approaches aggregate the local features as evidence to infer the global model parameters with the inherent human body structure as constraints. So if a joint is occluded, it can still rely on other joints to guess the possible locations that are consistent with the prior. However, the model-based approaches get larger $3$D errors than the model-free approaches due to the difficult optimization problems.

The second class is model-free approaches \citep{qiu2019cross,iskakov2019learnable,burenius20133D,PavlakosZDD17,dong2019fast,amin2013multi,belagiannis20143d,wang2019meta} which often follow a two-step framework. They first detect $2$D poses in images of all camera views. Then with the aid of camera parameters, they recover the $3$D pose using either triangulation \citep{amin2013multi,iskakov2019learnable} or pictorial structure models \citep{burenius20133D,PavlakosZDD17,dong2019fast}. Recursive pictorial structure model is introduced in \citep{qiu2019cross} to speed up the inference process. The authors in \citep{iskakov2019learnable} also propose to use learnable triangulation \citep{hartley2003multiple} for human pose estimation which is more robust to inaccurate $2$D poses. If the $2$D poses are accurate, the recovered $3$D poses are guaranteed to be accurate without worrying about being trapped in local optimum as the model-based methods. 

The development of more powerful network architectures \citep{newell2016stacked,sun2019deep} has dramatically improved the $2$D pose estimation accuracy on benchmark datasets, which in turn also decreases the $3$D pose estimation error. For example, on the most popular benchmark Human3.6M \citep{ionescu2014human3}, the $3$D MPJPE error has decreased to about $20$mm which can meet the requirements of many real-life applications.

\subsection{Improving ``In the Wild'' Performance}
\label{sec:related_wild}

\paragraph{Sensors} Occlusion is probably the biggest challenge for in-the-wild scenarios. One straightforward solution is to use additional sensors such as IMUs \citep{trumble2017total} and radio signals \citep{zhao2019through}, which are not impacted by occlusion. For example, Roetenberg \etal \citep{roetenberg2009xsens} place $17$ IMUs at the rigid bones. If the measurements are accurate, the $3$D pose is fully determined. In practice, however, the accuracy is limited by the drifting problem. To that end, some approaches \citep{trumble2017total,von2018recovering,gilbert2019fusing,malleson2017real,zhe2020fusingimu} propose to fuse images and IMUs to achieve more robust pose estimation. Some works \citep{zhao2019through,li2019making,zhao2018through} leverage the fact that wireless signals in the WiFi frequencies traverse walls and reflect off the human body, and propose a radio-based system that can estimate $2$D poses even when persons are completely occluded by walls. However, these approaches also have their own problems. For example, how to effectively fuse visual and inertial signals for IMU-based approaches? Besides, wearing sensors on the body is intrusive, and is not acceptable in some scenarios such as football games. On the other hand, the WiFi-based solutions cannot deal with self-occlusion which is a big limitation.

\paragraph{Data Augmentation} Collecting more images for model training is an effective approach to improve the generalization performance. For example in \citep{zhou2017towards,qiu2019cross}, the authors propose to use the MPII \citep{andriluka14cvpr} and the COCO \citep{lin2014microsoft} datasets to help train the $2$D module of the $3$D pose estimators which effectively reduces the risk of over-fitting to simple training datasets. However, annotating a sufficiently large pose dataset is expensive and time consuming. So some approaches \citep{rogez2016mocap,varol2017learning,hoffmann2019learning,chen2016synthesizing,lassner2017unite} propose to generate synthetic images. The main issue is to bridge the gap between the synthetic and real images such that the model trained on synthetic images can be applied to real images. To that end, some approaches such as \citep{peng2018jointly} propose to use generative adversarial networks to generate realistic images.

\paragraph{Spatial-Temporal Context Models} Some approaches propose to use spatial-temporal context models to jointly detect all joints in a video sequence such that each joint can benefit from other joints in the same or neighboring frames. Intuitively, if a body joint is occluded thus is difficult to be detected according to its own appearance, they can use the locations of other joints to guess the possible location. For example, in a previous work \citep{cao2017realtime,kreiss2019pifpaf}, the authors propose to detect body parts, \ie the links connecting two joints, in addition to the individual joints. This provides a chance to mutually enhance the detection of the two linked joints. In \citep{cheng2019occlusion,pavllo20193d}, temporal convolution is utilized to deal with occlusion in current frames. Some works such as \citep{qiu2019cross} propose to establish the spatial correspondence across multiple camera views, and leverage multi-view features for robust joint detection. Significant performance improvement has been achieved for the occluded joints on several benchmark datasets. The main drawback of the approach \citep{qiu2019cross} is the lack of flexibility in practice since it needs to train a separate fusion network for every possible camera placement. Our work differs from \citep{qiu2019cross} in that it can be applied to new environments with different numbers of cameras and different camera poses without additional adaptation. We will compare the two methods in the experiments.

\subsection{Consensus Learning}
\label{sec:related_agreement}
A fundamental problem in multi-sensor fusion is to detect and remove outliers as the sensors may produce inconsistent measurements. RANSAC \citep{fischler1981random} is the most commonly used outlier detection method. The main assumption is that the dataset consists of inliers. It produces reasonable results only with a certain probability which increases as the number of inliers. In practice, when the number of sensors is small, the probability of detecting the real outliers is also small. For example, in multiview human pose estimation, the number of cameras is only four to eight for most benchmark datasets \citep{ionescu2014human3,trumble2017total}. For such cases, we observe that RANSAC may not be the best option.

In recent years, uncertainty learning \citep{kendall2017uncertainties,gal2015dropout,lakshminarayanan2017simple,zafar2019face,lakshminarayanan2017simple,pleiss2017fairness} has attracted a lot of attention which is particularly important for high-risk applications such as autonomous driving and medical diagnosis \citep{gal2016uncertainty,ghahramani2016history}. The main idea is that, when a model makes a prediction, it also outputs a score reflecting the confidence of the prediction. Consider an autonomous car that uses a neural network to detect people. If the network is not confident about the prediction, the car could probably rely on other sensors for making the correct decision.
Uncertainty is introduced to computer vision in \citep{kendall2017uncertainties,kreiss2019pifpaf,he2019bounding,ilg2018uncertainty}. Another branch of approaches such as \citep{guo2017calibration,pleiss2017fairness} propose to learn uncertainty by calibration. They propose to train the model such that the probability associated with the predicted class label agrees with its ground truth correctness likelihood. 

The concept of uncertainty can be leveraged to reduce the impact of outliers. For example, in \citep{iskakov2019learnable}, the authors propose to predict an uncertainty score for each joint in each view. The score is used to weigh each view when doing triangulation. This dramatically reduces the $3$D pose estimation error. Inspired by the success of uncertainty learning in computer vision tasks, we propose to learn uncertainty for multiview feature fusion. The predicted uncertainty is used as a weight when fusing multiview features.  We show this adaptive feature fusion could effectively improve the fusion quality.

\section{The Basics for Multiview Fusion}
\label{sec:efl}

We first introduce the basics for multiview fusion to lay the groundwork for \emph{AdaFuse}. In particular, we discuss how to establish the point-point correspondence between two views such that the features correspond to the same $3$D space point can be fused together. The narrow baseline correspondence can be solved efficiently by local feature matching. However, in the context of multiview human pose estimation where only a small number of cameras are placed far away from each other, the local features cannot be robustly detected and matched especially for texture-less human regions. This poses a serious challenge.

To solve the problem, we present a coarse-to-fine approach to find matched points. It first establishes the point-to-line correspondence between two views by epipolar geometry, and then implicitly determine the point-to-point correspondence by exploring the sparsity of the heatmap representations. The approach notably simplifies the task because it avoids the challenging step of finding the exact correspondence. We first introduce epipolar geometry in section \ref{sec:epipolar} in order to determine the point-to-line correspondence. Then in section \ref{sec:heatmap}, we describe how we adapt epipolar geometry to perform multiview heatmap fusion. Finally, we discuss the side effect caused by the simplified fusion strategy and our solution in section \ref{sec:side}.

\begin{figure}
    \centering
    \includegraphics[width=0.8\linewidth]{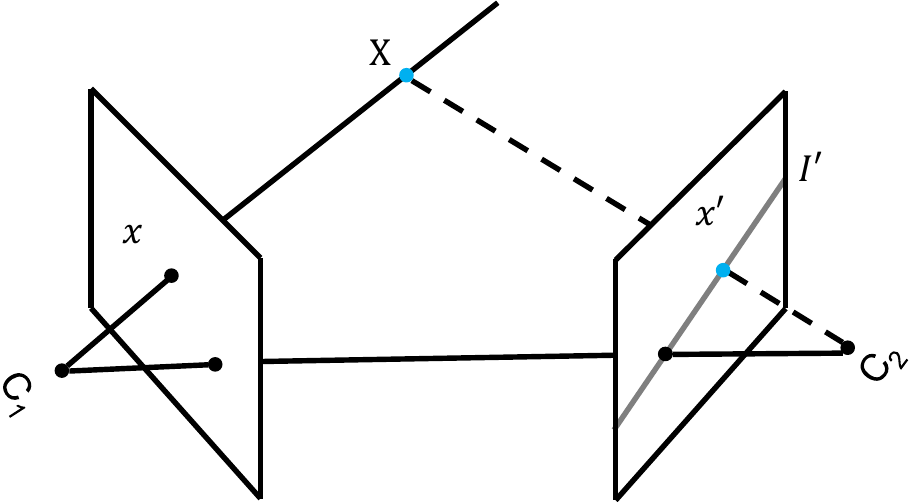}
    \caption{Illustration of the point-line correspondence in two views. For an arbitrary point $\mathbf{x}$ in one view, the corresponding point $\mathbf{x}'$ in another view has to lie on the epipolar line $\mathbf{I}^{\prime}$. This is the core of \emph{AdaFuse} for finding corresponding points in other views.}
    \label{fig:epipolar}
\end{figure}

\subsection{Epipolar Geometry}
\label{sec:epipolar}
{
Let us denote a point in $3$D space as $\mathbf{X} \in \mathcal{R}^{4\times 1}$ as shown in Figure \ref{fig:epipolar}. This could be the location of a body joint in the context of pose estimation. Note that homogeneous coordinate and column vector are used to represent a point. The $3$D point is imaged in two camera views, at $\mathbf{x}= \mathbf{P}\mathbf{{X}}$ in the first, and $\mathbf{x}' = \mathbf{P}'\mathbf{{X}} $ in the second, where $\mathbf{x}$ and $\mathbf{x}'  \in \mathcal{R}^{3 \times 1}$ represent $2$D points in images, $\mathbf{P}$ and $\mathbf{P}' \in \mathcal{R}^{3 \times 4}$ are the projection matrix for each camera.
Since the two $2$D points correspond to the same $3$D point and have the same semantic meanings, their features can be safely fused such that each view benefits from the other view.}

{
The epipolar geometry \citep{hartley2003multiple} between two views is essentially the geometry of the intersection of the image planes with the pencil of planes having the baseline as axis. The baseline is the line joining the camera centers $C_1$ and $C_2$.  In particular, for each location $\mathbf{x}$ in the first view, it helps us to determine the location of the corresponding point $\mathbf{x}'$ in the second view without having to know $\mathbf{X}$. }

{
We can see from Figure \ref{fig:epipolar} that the image points $\mathbf{x}$ and $\mathbf{x}'$, the $3$D point $\mathbf{X}$, and the camera centers $\mathbf{C_1}$ and $\mathbf{C_2}$ lie on the same plane $\mathbf{\pi}$. The plane intersects with the two image planes at epipolar lines $\mathbf{I}$ and $\mathbf{I}'$, respectively. In particular, }
\begin{equation}
\label{eq:epipolar}
\begin{split}
    &\mathbf{I}^{\prime} = \mathbf{F} \mathbf{x} \\
    &\mathbf{I} = \mathbf{F}^{\top} \mathbf{x}^{\prime},
\end{split}
\end{equation}
{
where $\mathbf{F} \in \mathcal{R}^{3\times 3}$ is fundamental matrix which can be derived from $\mathbf{P}$ and $\mathbf{P}'$. Readers can refer to \citep{hartley2003multiple} for detail derivation.}
In addition, the rays back-projected from $\mathbf{x}$ and $\mathbf{x}'$ intersect at $\mathbf{X}$, and the rays are coplanar, lying in $\mathbf{\pi}$. It is straightforward to derive that the location of $\mathbf{x}'$ which corresponds to $\mathbf{x}$ is guaranteed to lie on the epipolar line $\mathbf{I}^{\prime}$. However, we have to leverage additional information such as appearance to determine the exact location of $\mathbf{x}'$ on $\mathbf{I}^{\prime}$.

\begin{figure}
    \centering
    \includegraphics[width=0.98\linewidth]{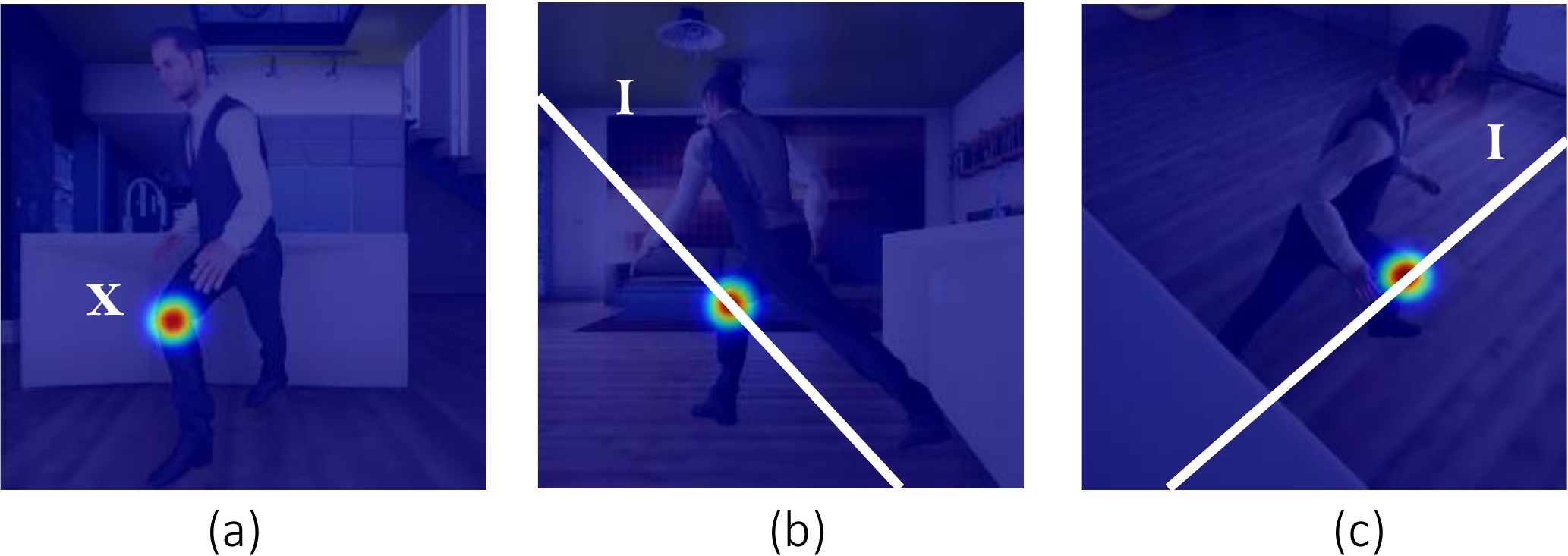}
    \caption{Epipolar geometry based heatmap fusion. For each location $\mathbf{x}$ in the first view, we first compute the corresponding epipolar lines in the other two views. Then we find the largest responses on the two lines, respectively and add them to the original response at $\mathbf{x}$.}
    \label{fig:heatmap}
\end{figure}

In the context of multiview feature fusion, for every image point $\mathbf{x}$, we need to find the corresponding point $\mathbf{x}'$ in the second view so that we can fuse the features at $\mathbf{x}$ with those at $\mathbf{x}'$ and obtain more robust pose estimations. Since we do not know the depth of $\mathbf{X}$, it could move freely on the line defined by the camera center $\mathbf{C_1}$ and image point $\mathbf{x}$.  However, we know that $\mathbf{x}'$ cannot span the entire image plane but is restricted to the line $\mathbf{I}^{\prime}$. In the following section 3.2, we will describe how we perform multiview feature fusion based on epipolar geometry.

{
\paragraph{Sampson Distance} 
In practice, usually we have 2D measurements $\mathbf{x}$ and $\mathbf{x}^\prime$ corresponding to the same 3D location $\mathbf{X}$ which is unknown. Due to measurement noise and errors, the line $\mathbf{C_1x}$ and $\mathbf{C_2x^\prime}$ might not intersect exactly at location $\mathbf{X}$. 
To obtain the optimal estimation for $\mathbf{X}$, we search for $\mathbf{\hat{X}}$ subject to }
\begin{equation}
\label{eq:reproj_error}
    d_{Reproj}^{2} = \min_{\mathbf{\hat{X}}} d^{2}\left(\mathbf{x}, \mathbf{P}\mathbf{\hat{X}}\right)+d^{2}\left(\mathbf{x}^{\prime}, \mathbf{P}^{\prime}\mathbf{\hat{X}}\right),
\end{equation}
{where
$d(\cdot)$ denotes Euclidean distance, $d_{Reproj}$ represents the reprojection distance \label{eq:reproj_error} between $\mathbf{x}$ and $\mathbf{x}^\prime$.} 
{Since there is optimization process when obtaining $d_{Reproj}$, we adopt an one-step method which is its first-order approximation \citep{hartley2003multiple}. This approximation is also called Sampson distance as}
\begin{equation}
\label{eq:sampson_distance}
    d_{Sampson} = \frac{\mathbf{x}^{\prime \top} \mathbf{F} \mathbf{x}}{(\mathbf{F} \mathbf{x})_{1}^{2}+(\mathbf{F} \mathbf{x})_{2}^{2}+\left(\mathbf{F}^{\top} \mathbf{x}^{\prime}\right)_{1}^{2}+\left(\mathbf{F}^{\top} \mathbf{x}^{\prime}\right)_{2}^{2}},
\end{equation}
{where $\mathbf{F}$ is fundamental matrix, the subscript $1$ or $2$ denotes the first or second element of a vector. By using Sampson distance, we can directly obtain distance between a pair of locations without knowing the intermediate $\mathbf{\hat{X}}$. In \emph{AdaFuse}, we use Sampson distance to represent to what extent a pair of $2$D joint detections support each other.}

\subsection{Heatmap Fusion}
\label{sec:heatmap}
Multiview fusion is applied to heatmaps rather than intermediate features as shown in Figure \ref{fig:overview}. This is because heatmap has the nice property of sparsity which can simplify the point-point matching. A heatmap produces a per-pixel likelihood for joint locations in the image. Specifically, it is generated as a two-dimensional Gaussian distribution centered at the coordinate of the joint. So it has a small number of large responses near the joint location, and a large number of zeros at other locations. See Figure \ref{fig:heatmap} (a) for an example heatmap of the right knee joint.

The sparse heatmaps allow us to safely skip the exact point-point matching because the features at the ``zero'' locations on the epipolar line are not contributing to the feature fusion. As a result, instead of trying to find the exact corresponding location in the other view, \emph{we simply select the largest response on the epipolar line as the matched point}. This is a reasonable simplification because the corresponding point usually has the largest response. For example, in Figure \ref{fig:heatmap}, for each location $\mathbf{x}$, we first compute the corresponding epipolar lines in the other two camera views. Then we find the largest responses on the two epipolar lines, respectively and fuse them with the response at $\mathbf{x}$.

Let us denote the heatmap in view $v$ as $\mathbf{H}^v$. The response at the location $\mathbf{x}$ of the heatmap is denoted as $\mathbf{H}^v(\mathbf{x})$. The corresponding epipolar line of $\mathbf{x}$ in view $u$ is denoted as $\mathbf{I}^u(\mathbf{x})$ which consists of a number of discrete locations on the heatmap $\mathbf{H}^u$. The epipolar line can be analytically computed based on the camera parameters for every location $\mathbf{x}$. Then we formulate multiview fusion as

\begin{equation}
\label{eq:fusion}
    \mathbf{\hat{H}}^v(\mathbf{x}) = \lambda \mathbf{H}^v(\mathbf{x}) + \frac{1-\lambda}{N} \sum_{u=1}^{N} \max_{\mathbf{x'} \in \mathbf{I}^u(\mathbf{x})}{\mathbf{H}^u(\mathbf{x}')},
\end{equation}
where $\mathbf{\hat{H}}$ denotes the fused heatmap and $N$ is the number of camera views which contribute to the fusion of current view. The parameter $\lambda$ balances the responses in the current and other views.

\begin{figure}
    \centering
    \includegraphics[width=0.8\linewidth]{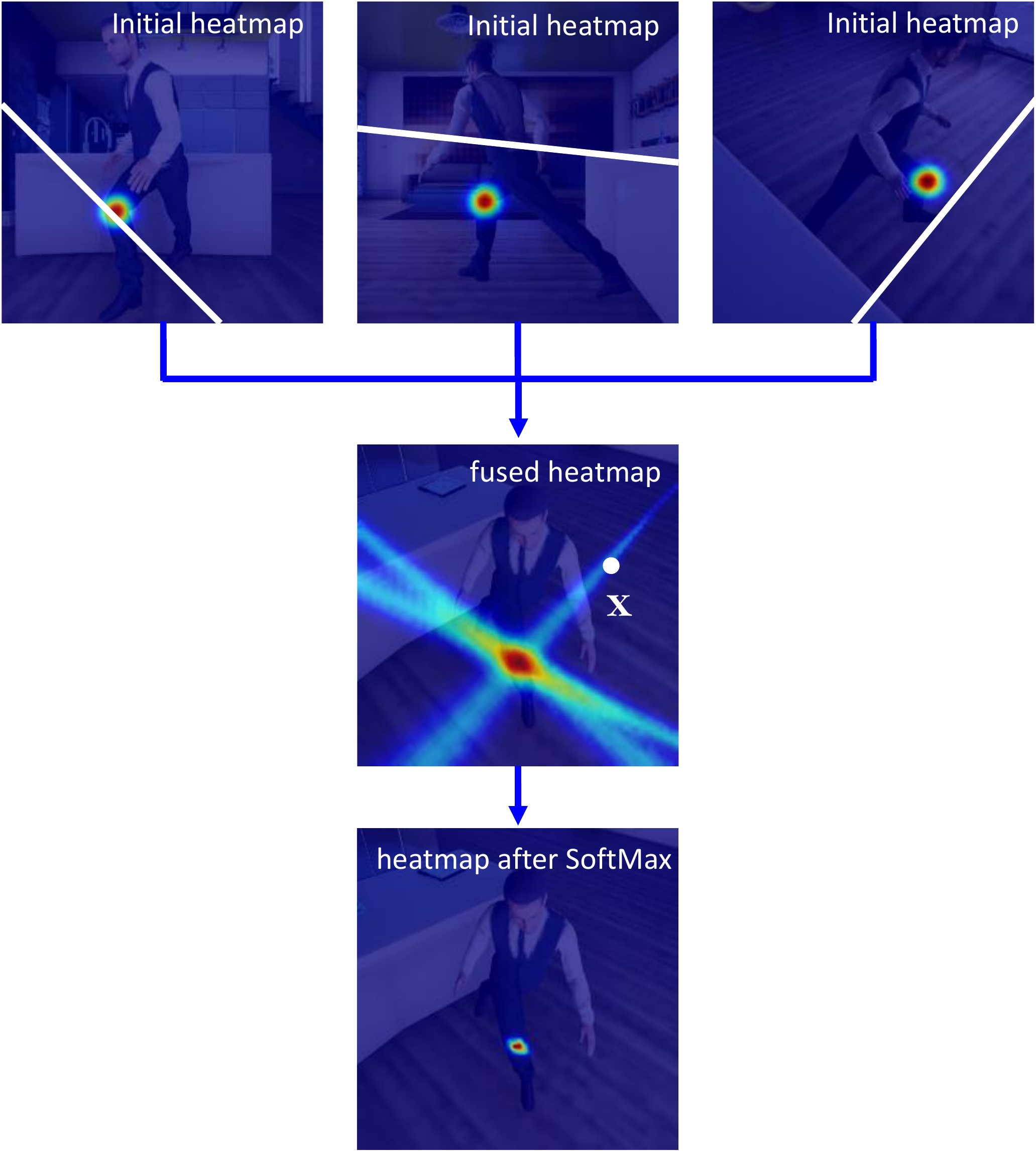}
    \caption{The ambiguity problem in our simplified multiview fusion approach and our solution. We can see from the ``fused heatmap'' that the correct location has the largest response which is as expected. However, for an incorrect location $\mathbf{x}$, there is also a chance that the response is also enhanced by at most one view. Fortunately, the correct location will be enhanced more times (three times in this example) leading to the largest response. So we apply the SoftMax operator to the fused heatmap to reduce the responses at incorrect locations.}
    \label{fig:ambiguity}
\end{figure}

\subsection{Side Effect and Solution}
\label{sec:side}
One side effect caused by the simplified fusion model (\ie Eq. (\ref{eq:fusion})) is that some background locations may be enhanced undesirably. We visualize an example in the second row of Figure \ref{fig:ambiguity}. We can see that many background pixels, for example $\mathbf{x}$, have non-zero responses which are caused by fusion. This phenomenon happens because multiple epipolar lines (in other views) may pass the ground truth joint location which has large responses, and some of the epipolar lines actually correspond to background pixels in the current view. This is explained in Figure \ref{fig:ambiguity}. For a location $\mathbf{x}$ in the current view, the corresponding epipolar lines in the other three views are drawn in the first row. We can see that although $\mathbf{x}$ is not at a meaningful joint location, the epipolar line in the first view passes the ground truth knee joint and leads to a large unexpected response for $\mathbf{x}$.

Fortunately, there are patterns for the background pixels that could be undesirably impacted. In general, the pixels that are impacted by a high response location in another view are guaranteed to lie on the same line. More importantly, the lines that correspond to different views do not overlap. It means, for a location $\mathbf{x}$ in the background, its response can only be enhanced by at most one view. In contrast, the location which corresponds to meaningful body joints will be enhanced by multiple views. In other words, the correct location is guaranteed to have the largest response for general cases. So we take advantage of this observation and directly apply the SoftMax operator to remove the small responses. See the third row in Figure \ref{fig:ambiguity} for the effect. We can see that only the large responses around the joint location are preserved.

\begin{figure}
    \centering
    \includegraphics[width=0.62\linewidth]{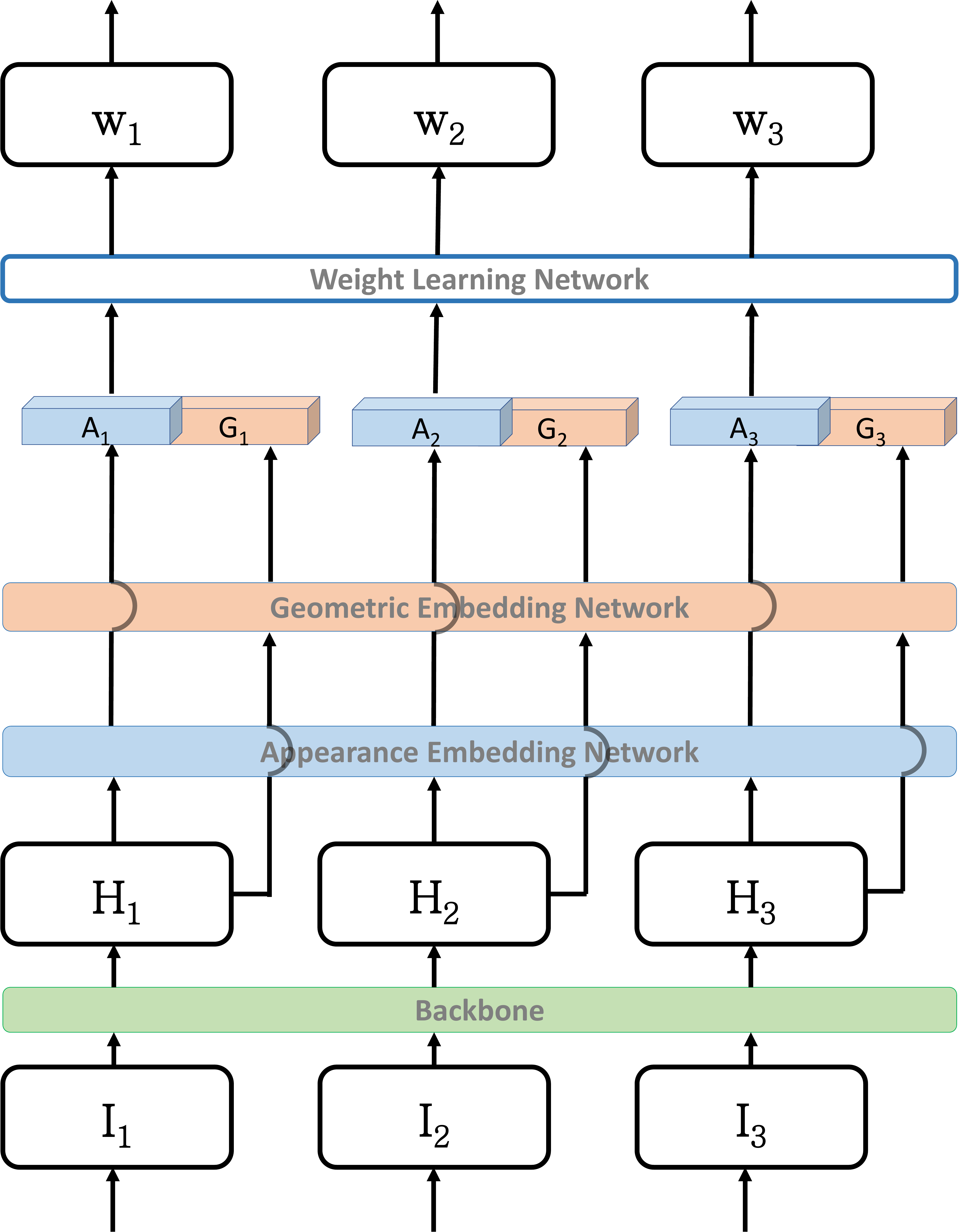}
    \caption{Network for learning adaptive fusion weights. The backbone network for pose estimation is used to extract heatmaps $\mathbf{H}_v$ for each view $\mathbf{I}_v$. The heatmaps are fed to \emph{appearance embedding network} and \emph{geometry embedding network}, respectively, to extract features, which are concatenated and fed to a \emph{weight learning network} to learn the fusion weights which reflect the heatmap quality in each view. The weights are used for multiview fusion. }
    \label{fig:weightlearning}
\end{figure}

\subsection{Implementation Details}
It is worth noting that the above fusion method does not have learnable parameters. So we only need to train the backbone network such as SimpleBaseline \citep{simplebaselines} to estimate pose heatmaps. The loss function for training the backbone network is defined as MSE loss between the estimated heatmaps and ground truth heatmaps. In the testing stage, given the heatmaps estimated by SimpleBaseline, we fuse them deterministically by our approach.

\section{Adaptive Weight for Multiview Fusion}
\label{sec:weightnet}
The fusion strategy introduced in the previous section treats all views evenly without considering the feature quality of each view. Note that the fusion weight is $\frac{1-\lambda}{N}$ for the $N$ views in Eq. (\ref{eq:fusion}). However, the strategy is problematic in some cases where the heatmaps of some camera views are incorrect. This is because those features may undesirably mess up the features in good views, leading to a completely incorrect $2$D pose estimation results.

To solve this problem, we present a weight learning network to learn an \emph{adaptive weight} for each view to faithfully reflect its heatmap quality. It takes inputs of the heatmaps of $N$-views extracted by the pose estimation network, and regresses $N$ weights $\omega^{u}$. Then multiview fusion is rewritten to consider the weights as follows 

\begin{equation}
\label{eq:adaptivefuse}
    \mathbf{\hat{H}}^v(\mathbf{x}) = \omega^v\mathbf{H}^v(\mathbf{x}) + \sum_{u=1}^{N} \omega^u \max_{\mathbf{x'} \in \mathbf{I}^u(\mathbf{x})}{\mathbf{H}^u(\mathbf{x}')},
\end{equation}

The prediction of the adaptive fusion weight $\omega$ is implemented by a lightweight neural network as shown in Figure \ref{fig:weightlearning}. On top of the heatmaps $\mathbf{H}$ provided by the pose estimation network, we extract two types of information for making the prediction. The first is the appearance embedding which extracts information such as the distribution characteristics of the heatmaps. The second is the geometry embedding which considers the cross-view location consistency. The two terms are complementary to each other. The proposed weight learning network can be joined with the pose estimation network for end-to-end training without enforcing supervision on the weights.

\subsection{The Appearance Embedding}
The heatmap of each joint actually contains rich information to infer its heatmap quality. For example, if the predicted heatmap has a desired shape of Gaussian kernel, then in many cases, the heatmap quality is good. In contrast, if the predicted heatmap has random and small responses all over the space (for example, when the joint is occluded), then the quality is likely to be bad.

\begin{figure}
    \centering
    \includegraphics[width=1\linewidth]{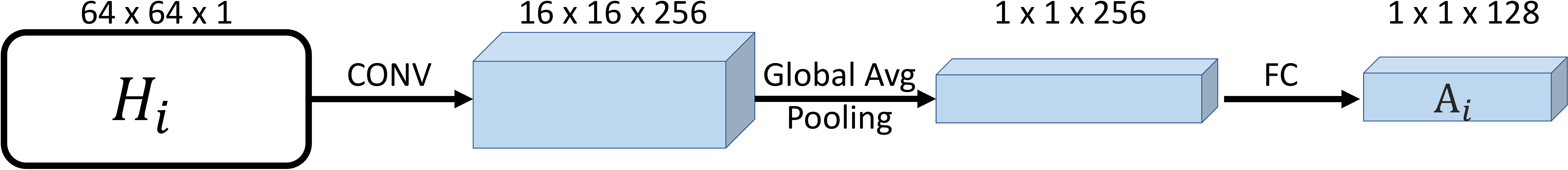}
    \caption{The appearance embedding network for predicting the fusion weight. 
    $i$ is the index of camera views. The parameters in the network are shared for all views and joints. See also Figure \ref{fig:weightlearning} for how the appearance embedding $A_i$ is used for determining the fusion weight. }
    \label{fig:appearance}
\end{figure}

We propose a simple network to extract appearance embeddings for each joint in each camera view. Figure \ref{fig:appearance} shows the network structure. Starting from the heatmaps $\mathbf{H}_i$, we apply a convolutional layer to extract features. Then the features are down-sampled by average pooling and fed to a Fully Connected (FC) layer for extracting the appearance embeddings. Different joint types and camera views share the same weights. We only show the network for a single view and a single joint for simplicity. The appearance embedding network is jointly learned end-to-end with the pose estimation network.

\begin{figure}[]
    \centering
    \includegraphics[width=1\linewidth]{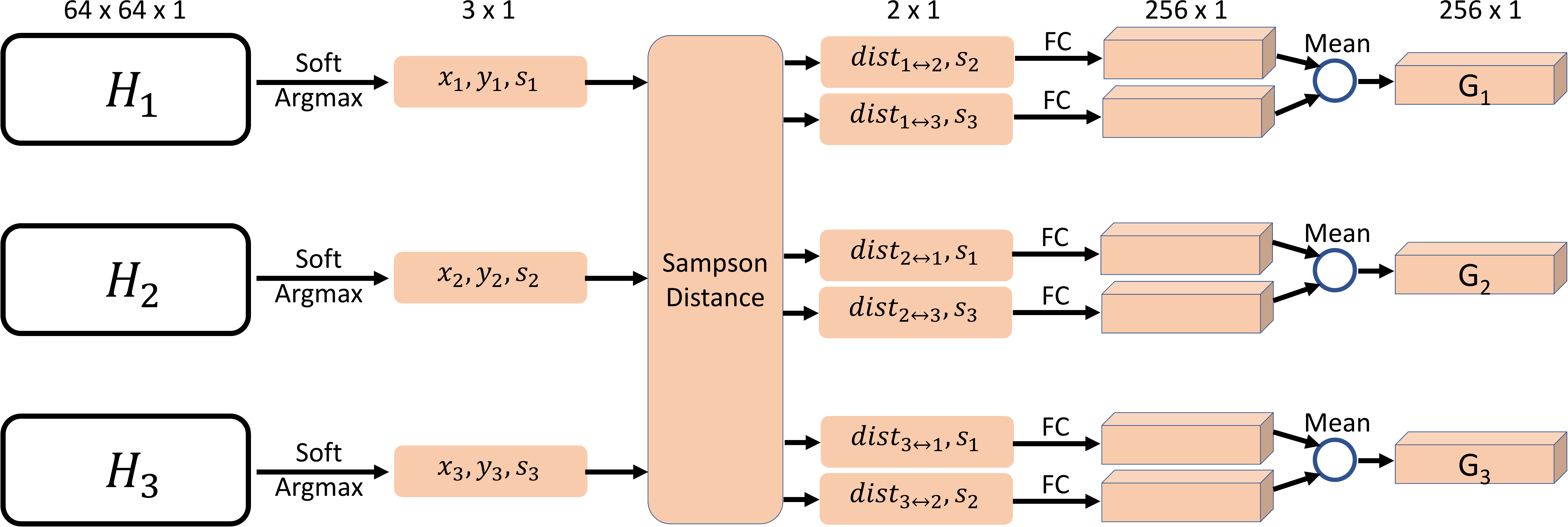}
    \caption{The geometry embedding network for predicting the fusion weight. For each joint in each camera view (three views are shown in this example), it generates a $256$-dimensional embedding to reflect the heatmap (pose) quality. Note that the FC is shared for all branches.}
    \label{fig:geometric}
\end{figure}

\subsection{The Geometry Embedding}
The appearance embedding alone is not sufficient for some challenging cases where the heatmaps have the desired shape of Gaussian kernel but at the wrong locations. One such example is when the left knee is detected at the location of right knee which is usually known as the ``double counting'' problem to the community. To solve this problem, we propose to leverage the location consistency information among all camera views. Our core motivation is that the predicted joint location in one camera view is more reliable if it agrees with the locations in other views.

\begin{figure}[hb]
    \centering
    \includegraphics[width=0.7\linewidth]{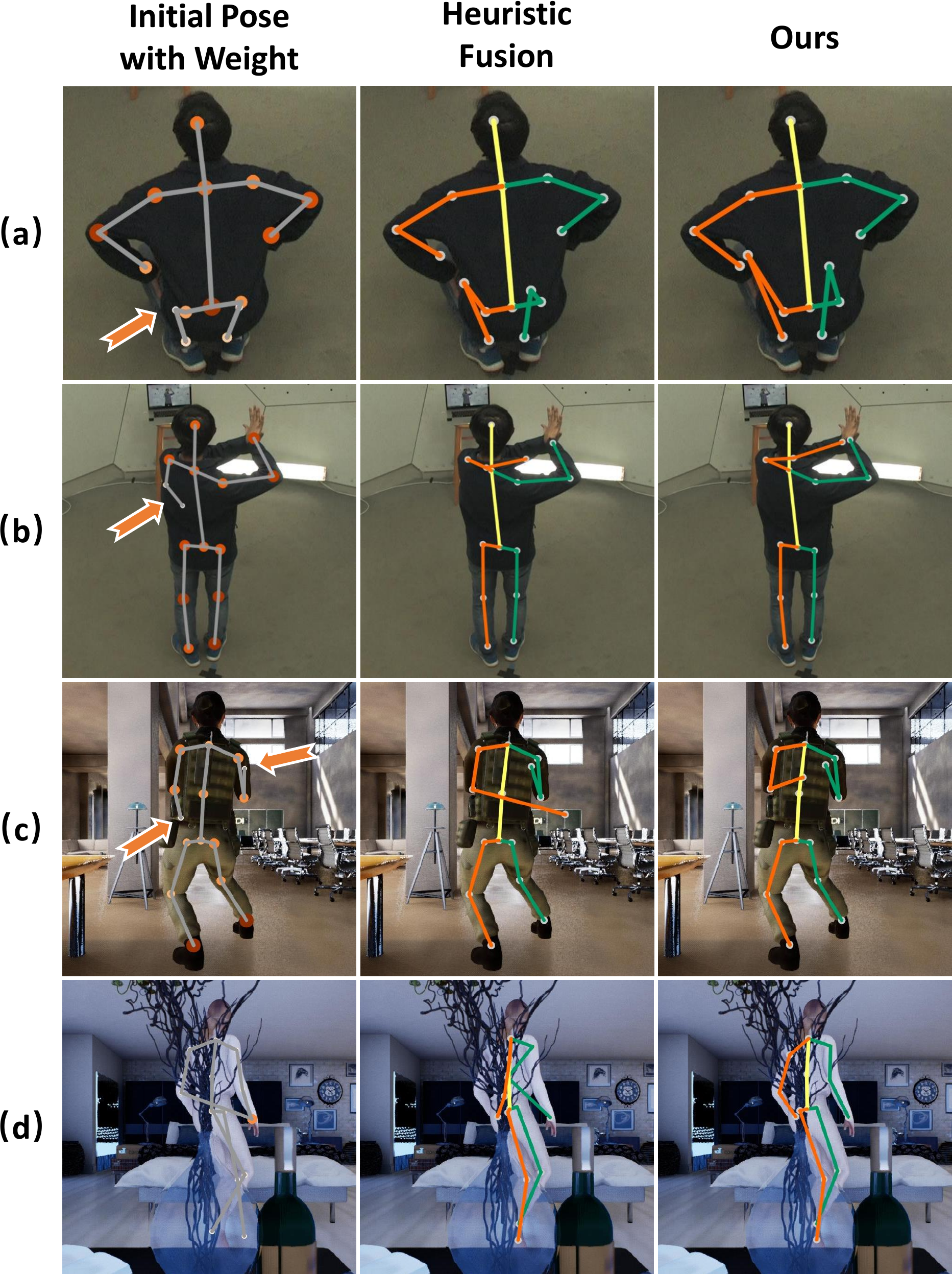}
    \caption{We visualize the predicted fusion weights by the size of the markers in the first column. A large marker denotes a larger weight. The rest two columns show the  poses estimated by \emph{HeuristicFuse} and \emph{AdaFuse}, respectively. Our \emph{AdaFuse} has clearly better estimations due to the consideration of the feature quality in every view.}
    \label{fig:weights}
\end{figure}

We implement this idea by a geometry embedding network as shown in Figure \ref{fig:geometric}. Starting from the heatmaps $\mathbf{H}$, we first apply the ``soft-argmax'' operator \citep{sun2018integral} to obtain the location $(x, y)$ of the joint in each view. We also get the heatmap response value $s$ in that location to reflect its confidence. Then we compute the Sampson distance \citep{hartley2003multiple} $dist_{i \leftrightarrow j}$ between the current view and other views to measure the correspondence or consistency error. A small $dist_{i \leftrightarrow j}$ means the joint locations in the two views are consistent. Intuitively, the location that is consistent with most views is more reliable. Finally, we propose to use a FC layer to embed the Sampson distance into a feature vector. The feature vectors of all camera pairs are then averaged to obtain the final geometry embedding.

\subsection{Weight Learning Network}
We propose a simple network consisting of three FC layers to transform the concatenated appearance and geometric embeddings to regress the final weight. It is worth noting that we do not train the weight learning network independently. Instead, we join it with the pose estimation network to minimize the fused $2$D heatmap loss without enforcing intermediate supervision on the fusion weights. The first column in Figure \ref{fig:weights} shows some example weights predicted by our approach. We can see that when the joints are occluded, and are localized at incorrect locations, the corresponding fusion weights are indeed smaller than other joints.

\begin{figure}
    \centering
    \includegraphics[width=1\linewidth]{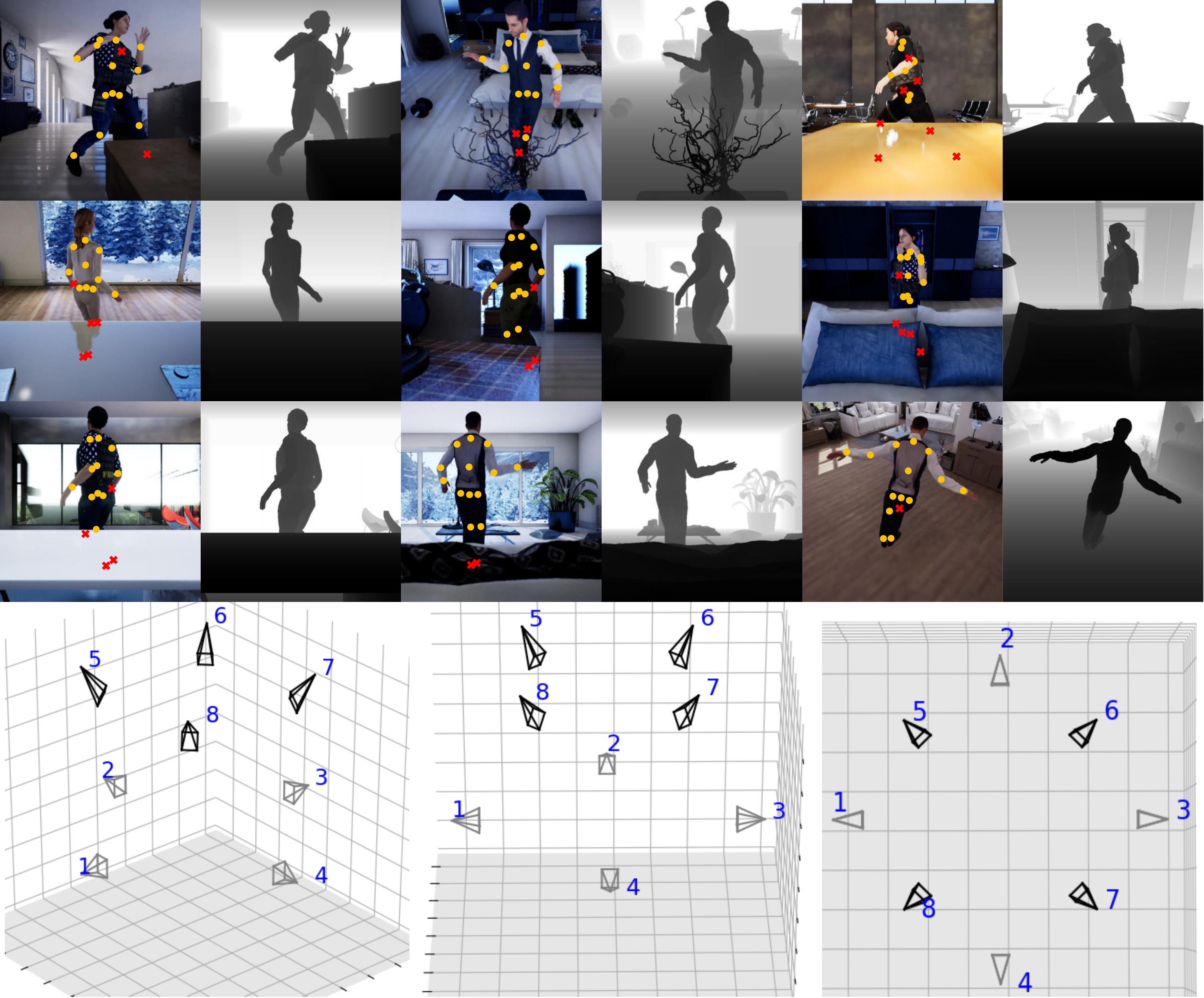}
    \caption{We show some typical images, ground-truth $2$D joint locations and the depth maps from the \emph{Occlusion-Person} dataset. The joint represented by red ``x'' means it is occluded. The \textbf{bottom} row shows spacial configuration of the eight cameras used in the dataset from different view angles.}
    \label{fig:occ-dataset}
\end{figure}

\section{Datasets and Metrics}
\label{sec:dataset}
We introduce the three datasets used for evaluation and the corresponding metrics. We also describe how we construct the synthetic person dataset \emph{Occlusion-Person} which has a large amount of human-object occlusion.

\begin{table}[h]
\centering
\caption{The statistics of the public multiview pose estimation datasets. Only the  \emph{Occlusion-Person} dataset provides occlusion labels.}
\label{table:unreal_statis}
\begin{tabular}{lcccc}
\toprule
Dataset          & Frames  & Cameras  & Occluded Joints \\ \hline
Human3.6M        & $784$k & 4       & - \\
Total Capture    & $236$k  & 8       & - \\
Panoptic         & $36$k  & 31        & - \\
Occlusion-Person & $73$k     & 8       & 20.3\%  \\
\toprule
\end{tabular}
\end{table}

\subsection{Datasets}
\paragraph{The Human3.6M Dataset \citep{ionescu2014human3}} It provides synchronized images captured by four cameras. There are seven subjects performing daily actions. We use a cross-subject evaluation scheme where subjects $1,5,6,7,8$ are used for training and $9,11$ for testing. We also use the MPII dataset \citep{andriluka14cvpr} to augment the training data to avoid over-fitting to the simple background. Since the MPII dataset provides only monocular images, we only train the backbone network before multiview fusion.

\paragraph{The Total Capture Dataset \citep{trumble2017total}} It provides synchronized person images captured by eight cameras. Following the dataset convention, the training set consists of ``ROM1,2,3", ``Freestyle1,2", ``Walking1,3", ``Acting1,2" and ``Running1"   on subjects 1, 2 and 3. The testing set consists of ``Freestyle3 (\textbf{FS3})", ``Acting3 (\textbf{A3})" and ``Walking2 (\textbf{W2})" on subjects 1,2,3,4 and 5.

\paragraph{The CMU Panoptic Dataset \citep{joo2019panoptic}}
This recently introduced dataset provides images captured by dozens of cameras. We uniformly select six cameras to evaluate the impact of the number of cameras on $3$D pose estimation. In particular, the cameras 1, 2, and 10 are firstly selected to construct a 3-view experiment setting. Then the cameras 13, 3 and 23 are sequentially added to the previous three cameras to construct a four, five and six view experiment setting, respectively. We follow the practice of the previous work \citep{xiang2019monocular} to select the training and testing sequences which consist of only one person. Since few works have reported numerical results on this dataset, we only compare our approach to the baselines.

\paragraph{The Occlusion-Person Dataset}
The previous benchmarks do not provide occlusion labels for the joints in images which prevents us from performing numerical evaluation on the occluded joints. In addition, the amount of occlusion in the benchmarks is limited. To address the limitations, we propose to construct this synthetic dataset \emph{Occlusion-Person}. We adopt UnrealCV \citep{qiu2017unrealcv} to render multiview images and depth maps from $3$D models. In particular, thirteen human models of different clothes are put into nine different scenes such as living rooms, bedrooms and offices. The human models are driven by the poses selected from the CMU Motion Capture database. We purposely use objects such as sofas and desks to occlude some body joints. Eight cameras are placed in each scene to render the multiview images and the depth maps. {The eight cameras are placed evenly  
every 45 degree on a circle of two meters radius at about $0.9$ and $2.3$ meters high, respectively.} We provide the $3$D locations of 15 joints as ground truth. Figure \ref{fig:occ-dataset} shows some sample images from the dataset and spacial configuration of the cameras.

The occlusion label for each joint in an image is obtained by comparing its depth value (available in the depth map), to the depth of the $3$D joint in the camera coordinate system. If the difference between the two depth values is smaller than $30$cm, then the joint is not occluded. Otherwise, it is occluded.  Table \ref{table:unreal_statis} compares this dataset to the existing benchmarks. In particular, about $20\%$ of the body joints are occluded in our dataset. We use $75\%$ of the dataset for training and $25\%$ for validation.

\begin{table*}[ht]
\centering
\caption{The $2$D pose estimation accuracy (PCKh@t) of the baseline methods and our approach on the Human3.6M dataset. We report results for each individual joint and the average over all joints. }
\label{table:pck_h36m}
\begin{tabular}{l|ccccccccccc|c}
\toprule
  Methods    & Root  & Belly & Neck  & Nose  & Head  & Hip  & Knee  & Ankle  & Shlder   & Elbow   & Wrist  & \emph{Mean}   \\ \hline
 NoFuse        & 95.8 & 77.1 & 60.4 & 86.4 & 86.2 & 79.3 & 81.5 & 58.6 & 65.1 & 78.3 & 70.1 & 74.8 \\
                     HeuristicFuse & 96.0 & 79.3 & 60.7 & \textbf{88.4} & \textbf{86.8} & 83.1 & 84.5 & 60.0 & \textbf{66.9} & 82.1 & 75.2 & 77.3 \\
                     ScoreFuse     & 96.2 & 79.3 & 61.6 & 88.3 & 86.2 & 83.3 & 84.3 & 60.5 & 66.6 & 83.1 & 77.4 & 77.8 \\
                     AdaFuse (Ours)       & \textbf{96.2} & \textbf{79.3} & \textbf{61.6} & 88.3 & 86.3 & \textbf{83.5} & \textbf{86.4} & \textbf{61.1} & 66.7 & \textbf{86.0} & \textbf{80.1} & \textbf{78.8} \\
\toprule
\end{tabular}
\end{table*}

\begin{table*}[ht]
\centering
\caption{The 3D pose estimation error ($mm$) of the baseline methods and our approach on the Human3.6M dataset. }
\label{table:ablation_h36m_mpjpe}
\begin{tabular}{l|cccccccccc|c}
\toprule
Methods      & Belly & Neck & Nose & Head & Hip  & Knee & Ankle  & Shlder  & Elbow  & Wrist  & Mean \\ \hline
NoFuse        & 21.6  & 16.8 & 15.7 & 11.3 & \textbf{17.8}     & 25.8     & 35.8     & 22.0     & 26.8     & 34.1     & 22.9  \\
HeuristicFuse      & 21.6  & 16.8 & 15.7 & 11.0 & 17.9     & 23.0     & 32.7     & 21.9     & 25.0     & 25.7     & 21.0  \\
ScoreFuse           & 21.4  & 16.7 & 15.8 & 10.9 & 18.3     & 21.3     & 30.8     & 21.8     & 23.3     & 23.2     & 20.1  \\
RANSAC         & 21.6  & 16.8 & \textbf{15.7} & 11.2 & 17.9     & 23.9     & 34.6     & 22.0     & 25.8     & 28.2     & 21.8  \\
AdaFuse (Ours)     & {\textbf{21.3}}  & \textbf{16.7} & 15.8 & \textbf{10.9} & 18.3     & \textbf{20.6}     & \textbf{30.2}     & \textbf{21.8}     & \textbf{21.3}     & \textbf{21.1}     & \textbf{19.5}  \\
\toprule
\end{tabular}
\end{table*}

\subsection{Metrics}

\paragraph{2D Metrics} The Percentage of Correct Keypoints (PCK) metric introduced in \cite{andriluka14cvpr} is commonly used for $2$D pose evaluation. PCKh@$t$ measures the percentage of the estimated joints whose distance from the ground-truth joints is smaller than $t$ times of the head length. Following the previous works, we report results when $t$ is $\frac{1}{2}$. Since the head length is not provided in the used three benchmarks, we approximately set it to be $2.5\%$ of the human bounding box width for all benchmarks.

\paragraph{3D Metrics}  The $3$D pose estimation accuracy is measured by Mean Per Joint Position Error (MPJPE) between a ground truth $3$D pose $y=[p^3_1,\cdots,p^3_M]$ and an estimated $3$D pose $\bar{y}=[\bar{p^3_1},\cdots,\bar{p^3_M}]$: $\text{MPJPE}=\frac{1}{M} \sum_{i=1}^M \|p^3_i-\bar{p^3_i}\|_2$ where $M$ is the number of joints in a pose.
We do not align the estimated $3$D poses to the ground truth by Procrustes. This is referred to as protocol 1 in some works \citep{martinez2017simple,tome2018rethinking}

\section{Experimental Results}
\label{sec:experiments}
We compare our approach to four baselines. The first is \emph{NoFuse} which estimates $2$D poses independently for each view without multiview fusion. The second is \emph{HeuristicFuse} which assigns a fixed fusion weight for each view according to Eq. (\ref{eq:fusion}). The parameter $\lambda$ is set to be $0.5$ by cross-validation. 
{The third baseline is \emph{ScoreFuse} which uses the same formulation as AdaFuse, i.e. Eq. (\ref{eq:adaptivefuse}), for feature fusion. It differs from AdaFuse only in the way we compute $\omega$. In particular, ScoreFuse computes $\omega$ as the maximum value of the heatmap $\mathbf{H}$.}
Our approach is denoted as \emph{AdaFuse} which uses the predicted weight for fusion as in Eq. (\ref{eq:adaptivefuse}). All of the four methods use triangulation \citep{hartley2003multiple} to estimate $3$D pose from the multiview $2$D poses.  We also compare to a baseline \emph{RANSAC} which does not perform multiview fusion, but uses RANSAC to remove the outliers in triangulation.

\subsection{Results on Human3.6M}
\paragraph{$2$D Pose Estimation Results}
The $2$D pose estimation results are presented in Table \ref{table:pck_h36m}. All multiview fusion methods remarkably outperform \emph{NoFuse}. The improvement is most significant for the Elbow and Wrist joints because they are frequently occluded by human body. The results demonstrate that multiview fusion is an effective strategy to handle occlusion. \emph{AdaFuse} achieves the highest {average} accuracy among all fusion methods validating that learning appropriate fusion weights can effectively reduce the negative impact caused by the features of low-quality views.

\begin{figure}[h]
    \centering
    \includegraphics[width=1\linewidth]{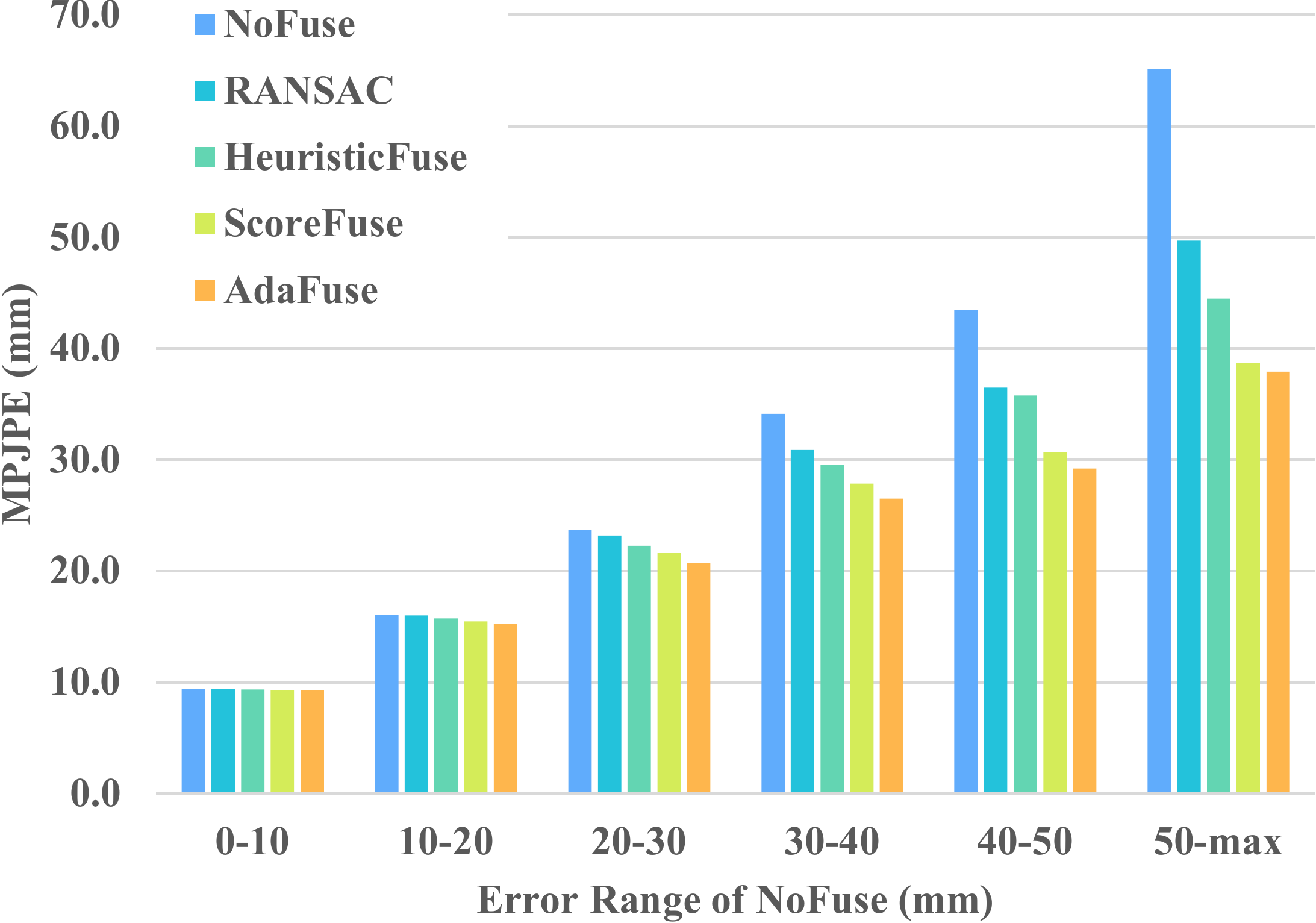}
    \caption{We divide the test set of Human3.6M into to six groups according to the error of \emph{NoFuse}. We compute the average error for every baseline and every group, respectively.}
    \label{fig:h36m_mpjpe_scale}
\end{figure}

\paragraph{$3$D Pose Estimation Results}
Table \ref{table:ablation_h36m_mpjpe} shows the $3$D pose estimation errors of the baselines and our approach. We can see that \emph{NoFuse} gets an average error of $22.9$mm. This is a very strong baseline whose error is only slightly larger than the state-of-the-arts (see Table \ref{table:mpjpe_h36m}). On top of this strong baseline, we observe that adding multiview fusion can further reduce the $3$D pose estimation errors.

\emph{HeuristicFuse} gets a smaller error than \emph{NoFuse} which is consistent with the $2$D results in Table \ref{table:pck_h36m}. The mean error only decreases by $1.9$mm because most examples are relatively easy leaving little space for improvement. However, significant improvement is achieved for the challenging joints such as Wrist. The \emph{ScoreFuse} gets a smaller error than \emph{HeuristicFuse}. It means assigning small weights to low-quality views helps improve the quality of the fused heatmaps. Finally, our approach \emph{AdaFuse}, which determines the fusion weight by considering both appearance cues and geometry consistency, notably decreases the average error to $19.5$mm. Considering the baseline is already very strong, the improvement is significant.
{We notice that AdaFuse achieves slightly worse results on a small number of joints such as hip and head. This is mainly because these joints are rarely occluded in the datasets so the $2$D pose estimator can obtain very accurate estimations for them. Further applying cross view fusion will introduce small noise to heatmaps leading to slightly worse $2$D pose estimation accuracy. But when occlusion occurs which is often the case in practice, the benefit brought by cross view fusion will be much more significant than the harm caused by the small noise.}

\emph{RANSAC} is the de facto standard for solving robust estimation problems. As shown in Table \ref{table:ablation_h36m_mpjpe}, it outperforms \emph{NoFuse} by removing some outlier $2$D poses in triangulation. However, it is not as effective as the multiview fusion methods because the latter also attempt to refine, in addition to removing, the outlier poses. Another reason is that the number of cameras in this task is small which reduces the chance of finding the true outliers. In addition, we find that \emph{RANSAC} is very sensitive to the threshold used for determining whether a data point is inlier or outlier. In our experiments, we set the threshold by cross validation.

\begin{figure}[ht]
    \centering
    \includegraphics[width=1\linewidth]{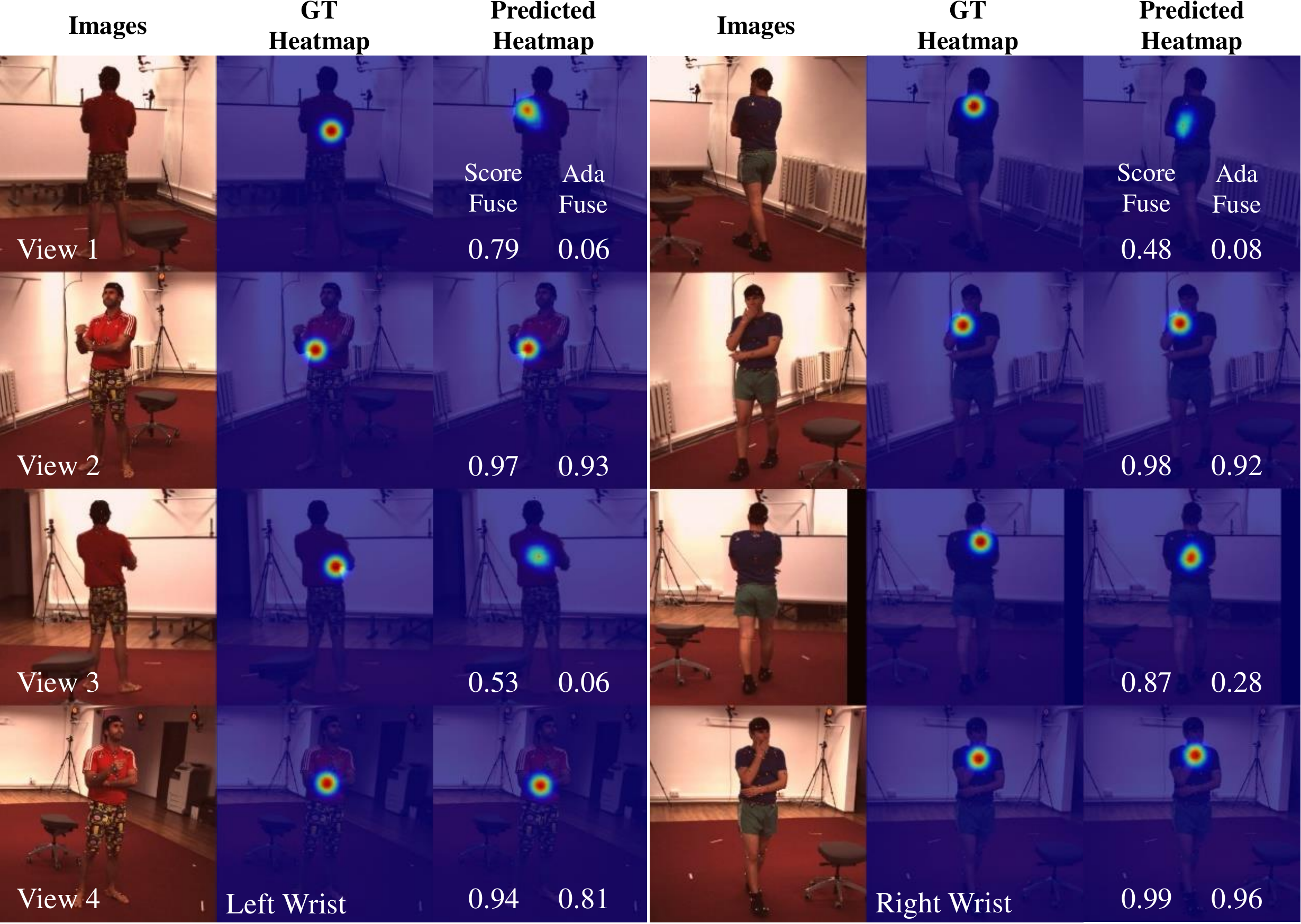}
    \caption{We visualize the weights predicted by the \emph{ScoreFuse} and \emph{AdaFuse}, respectively. For example, in the first example (left sub-figure), the pose estimation network generates a high response at the wrong location for the first view. Consequently, \emph{ScoreFuse} undesirably gives a large weight. In contrast, \emph{AdaFuse} gives a small weight by identifying that its location are inconsistent with other views.}
    \label{fig:h36m_maxv_ours}
\end{figure}

To better understand the improvement brought by \emph{AdaFuse}, we divide the testing samples of the Human3.6M dataset into six groups according to the $3$D errors of \emph{NoFuse}. Then we compute the average error for each group. Figure \ref{fig:h36m_mpjpe_scale} shows the results of various baselines. We can see that \emph{AdaFuse} achieves the most significant improvement when the original error of \emph{NoFuse} is large. However, even when the pose estimations of \emph{NoFuse} are already accurate, \emph{AdaFuse} can still reduce the error slightly.

\paragraph{Ablation Study on Fusion Weights}
One typical situation where \emph{ScoreFuse} fails is when the pose estimation network generates large scores at \emph{inaccurate} locations. In this case, \emph{AdaFuse} can outperform \emph{ScoreFuse} by leveraging the multiview geometry consistency. To support this conjecture, we visualize some typical heatmaps and the corresponding fusion weights predicted by the two methods, respectively, in Figure \ref{fig:h36m_maxv_ours}. We find that the heatmap responses are large for the four views although the locations are inaccurate for the first and third view. \emph{ScoreFuse} gives large weights for all views which finally leads to a corrupted heatmap. In contrast, \emph{AdaFuse} identifies that the predicted locations in the first and third view are inconsistent with the other two views in spite of their large scores. So it decreases the weights to ensure the good quality of the fused heatmap.

In addition, we also conduct ablation study on \emph{AdaFuse} by using only one of two embedding networks. When we only use either the \emph{appearance embedding} or \emph{geometry embedding}, the $3$D errors increase to $20.3$mm and $19.9$mm, respectively. Note that the improvement is actually much larger on those challenging examples. The results validate that the two embeddings are complementary.

\paragraph{Comparison to the State-of-the-arts}
Table \ref{table:mpjpe_h36m} compares our approach to the state-of-the-arts. We can see that our approach outperforms all of them. Note that two approaches, \ie \emph{Triangulation} and \emph{Volumetric}, are used in \citep{iskakov2019learnable} to lift $2$D poses to $3$D. The \emph{Triangulation} approach is more comparable to ours. Our approach \emph{AdaFuse} decreases the error of \citep{iskakov2019learnable} by about $13\% (= \frac{22.6-19.5}{22.6})$. The improvement is significant considering that the error of the state-of-the-art is already very small.

\begin{table*}[]
\small
\caption{The $3$D pose estimation errors ($mm$) of the state-of-the-arts and our approach on the Human3.6M dataset. We report results for each of the $15$ actions individually and also the average error over all actions. T- \cite{iskakov2019learnable} means triangulation is used. V- \cite{iskakov2019learnable} means volumetric method is used.
}
\label{table:mpjpe_h36m}
\setlength{\tabcolsep}{1mm}{
\begin{tabular}{l|ccccccccccccccc|c}
\toprule
             Methods & Direct & Disc. & Eat & Greet & Phone & Photo & Pose  & Purch & Sit & SitD & Smoke & Wait  & WalkD & Walk  & WalkT & \emph{MPJPE}  \\ \hline
\cite{trumble2017total}             &92.7 &85.9 &72.3 &93.2 &86.2 &101.2 &75.1 &78.0 &83.5 &94.8 &85.8 &82.0 &114.6 &94.9 &79.7 &87.3 \\
\cite{PavlakosZDD17} &41.2 &49.2 &42.8 &43.4 &55.6 &46.9 &40.3 &63.7 &97.6 &119.0 &52.1 &42.7 &51.9 &41.8 &39.4 &56.9 \\
\cite{tome2018rethinking} & 43.3 &49.6 &42.0 &48.8 &51.1 &64.3 &40.3 &43.3 &66.0 &95.2 &50.2 &52.2 &51.1 &43.9 &45.3 &52.8 \\
\cite{qiu2019cross} & 24.0 &26.7 &23.2 &24.3 &24.8 &22.8 &24.1 &28.6 &32.1 &26.9 &30.9 &25.6 &25.0 &28.0 &24.4 &26.2 \\
T- \cite{iskakov2019learnable}  & 20.4 &22.6 &20.5 &19.7 &22.1 &20.6 &19.5 &23.0 &25.8 &33.0 &23.0 &21.6 &20.7 &23.7 &21.3 &22.6 \\
V- \cite{iskakov2019learnable}  &18.8 &20.0 &19.3 &\textbf{18.7} &20.2 &19.3 &\textbf{18.7} &22.3 &\textbf{23.3} &29.1 &21.2 &\textbf{20.3} &19.3 &21.6 &19.8 &20.8\\

\hline
NoFuse        & 20.1 & 22.2 & 20.2 & 22.2 & 23.9 & 18.2 & 20.6 & 25.9 & 37.0 & 24.6 & 22.4 & 22.5 & 18.2 & 22.8 & 18.5 & 22.9 \\
AdaFuse (Ours)      & \textbf{17.8} & \textbf{19.5} & \textbf{17.6} & 20.7 & \textbf{19.3} & \textbf{16.8} & 18.9 & \textbf{20.2} & 25.7 & \textbf{20.1} & \textbf{19.2} & 20.5 & \textbf{17.2} & \textbf{20.5} & \textbf{17.3} & \textbf{19.5} \\

\toprule
\end{tabular}
}
\end{table*}

\subsection{Results on Panoptic}

\begin{table}[]
\scriptsize
\centering
\caption{The $2$D pose estimation accuracy (PCKh@t) of the baselines and our approach for the \textbf{occluded joints} on the \emph{Occlusion-Person} dataset. We report results for each joint type individually, and also the average accuracy over all joint types.}
\label{table:unreal_pckh_occluded}
\begin{tabular}{l|cccccc|c}
\toprule
 Methods       & Hip   & Knee  & Ankle  & Shlder   & Elbow   & Wrist & Avg.  \\ \hline

 NoFuse           & 63.4 & 21.5 & 17.0  & 29.5 & 14.6 & 12.4 & 30.9 \\
HeuristicFuse    & 76.9 & 59.0 & 73.4  & 63.5 & 49.0 & 54.8 & 65.0 \\
ScoreFuse        & 90.9 & 88.6 & 88.1  & 86.0 & 93.2 & 86.8 & 89.8 \\
 AdaFuse & \textbf{96.5} & \textbf{96.0} & \textbf{92.5}  & \textbf{94.1} & \textbf{98.3} & \textbf{93.2} & \textbf{95.5} \\
\toprule
\end{tabular}
\end{table}

\begin{figure}[]
    \centering
    \includegraphics[width=0.95\linewidth]{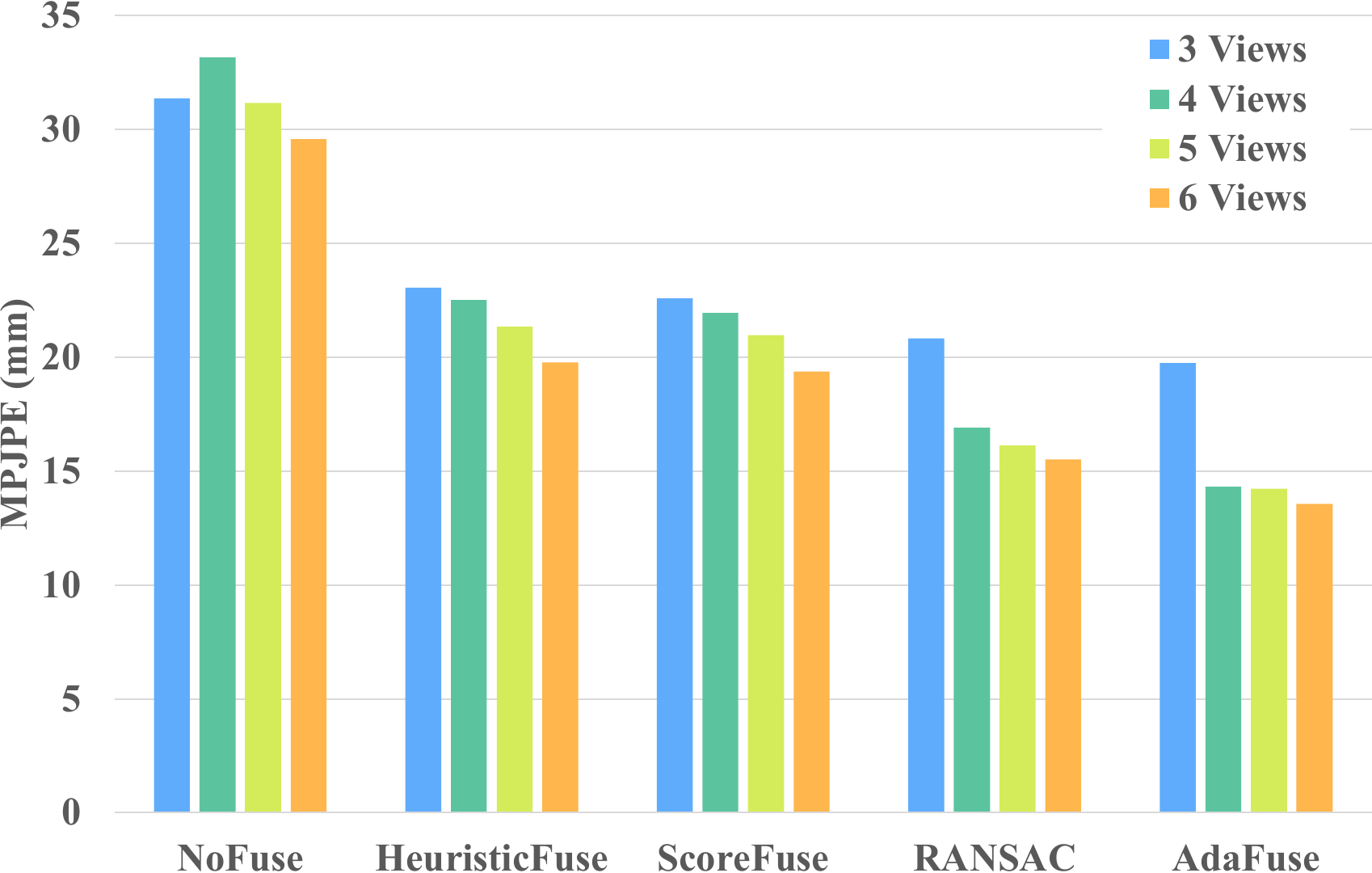}
    \caption{The $3$D pose estimation errors on the Panoptic dataset when different numbers of cameras are used. }
    \label{fig:popt_nview_avg}
\end{figure}

We evaluate the impact of the number of cameras on this dataset. Figure \ref{fig:popt_nview_avg} shows the mean $3$D errors when three to six cameras are used, respectively. In general, the error decreases when more cameras are used for most baselines. However, we observe that the error of \emph{NoFuse} actually becomes larger when the camera number increases from three to four. This undesirable phenomenon happens because the new camera view is very challenging thus the $2$D pose estimation results are inaccurate. However, for our approach \emph{AdaFuse}, the negative impact of low-quality heatmaps in individual views is limited due to the adaptive multiview fusion. We can see that the error of \emph{AdaFuse} consistently decreases when the number of cameras increases. Since there is not a commonly adopted evaluation protocol and very few works have reported results on this new dataset, we do not compare our approach to the other approaches.

\subsection{Results on Occlusion-Person}

\paragraph{$2$D Pose Estimation Results}
Table \ref{table:unreal_pckh_occluded} shows the results on the \emph{occluded} joints. Only about $30.9\%$ of the occluded joints can be accurately detected by \emph{NoFuse}. The result is reasonable because the features of the occluded joints are severely corrupted. All of the three multiview fusion methods remarkably improve the accuracy. In particular, more than $90\%$ of the occluded joints are correctly detected by \emph{AdaFuse}. The results demonstrate the advantages of our strategy for learning the fusion weights.

\paragraph{$3$D Pose Estimation Results}
We show the 3D pose estimation error ($mm$) for each joint type in Table \ref{table:unreal_mpjpe}. \emph{NoFuse} results in a large error of $48.1$mm. By improving the 2D pose estimation results on the occluded joints, the 3D errors are also significantly reduced, especially for the joints on the limbs such as Ankles and Wrists. In particular, our approach decreases the $3$D error significantly to $12.6$mm. 

\paragraph{Impact of Number of Occluded Views}
We also evaluate the impact of the number of occluded views on this dataset. In particular, we classify each joint into one of five groups according to the number of occluded views, and report the average joint error for each group, respectively. The results are shown in Table \ref{table:unreal_occluded_mpjpe}. We can see that when the joints are visible in all views, the simple baseline \emph{NoFuse} also achieves a very small error of $13.0$mm. However, the error increases dramatically to $82.6$mm when four views are occluded. Recall that there are eight views in total for this dataset. In contrast, the multiview fusion methods, especially our \emph{AdaFuse}, achieves consistently smaller errors than \emph{NoFuse}. More importantly, the error increase is much slower than \emph{NoFuse} when more camera views are occluded which validates the robustness of our approach to occlusion.

\begin{table*}[]
\centering
\caption{The 3D pose estimation error (\textit{mm}) of the baselines and our approach on the \emph{Occlusion-Person} dataset. We report the result on each joint individually and also the average over all joints. The second row shows the percentage of the joints that are occluded for each joint type.}
\label{table:unreal_mpjpe}
\begin{tabular}{l|ccccccccc|c}
\toprule
        & Root & Belly & Neck  & Hip   & Knee  & Ankle  & Shlder   & Elbow   & Wrist   & Mean \\ 
        Occluded (\%) & 14.3\%	&13.7\%	&7.6\%	&23.0\%	&25.0\%	&23.5\%	& 16.8\%	&25.3\%	&21.7\% &\\ \hline
NoFuse           & 10.0 & 12.2 & 12.5 & 16.8 & 61.1 & 113.9 & 28.0 & 63.7 & 60.3 & 48.1 \\
HeuristicFuse    & 8.8  & 10.7 & \textbf{11.5} & 14.2 & 21.1 & 19.2  & 17.5 & 23.6 & 24.1 & 18.0 \\
ScoreFuse        & 8.4  & 12.6 & 12.6 & 14.7 & 17.5 & 17.1  & 16.1 & 13.2 & 16.9 & 15.0 \\
RANSAC           & 8.6  & 11.2 & 11.7 & 12.9 & 18.8 & 17.9  & 17.1 & 14.5 & 19.7 & 15.5 \\

AdaFuse (Ours) & \textbf{7.2}  & \textbf{10.6} & \textbf{11.6} & \textbf{11.7} & \textbf{13.8} & \textbf{15.7}  & \textbf{14.2} & \textbf{9.9}  & \textbf{14.4} & \textbf{12.6} \\
\toprule
\end{tabular}
\end{table*}

\begin{figure*}
    \centering
    \includegraphics[width=0.85\linewidth]{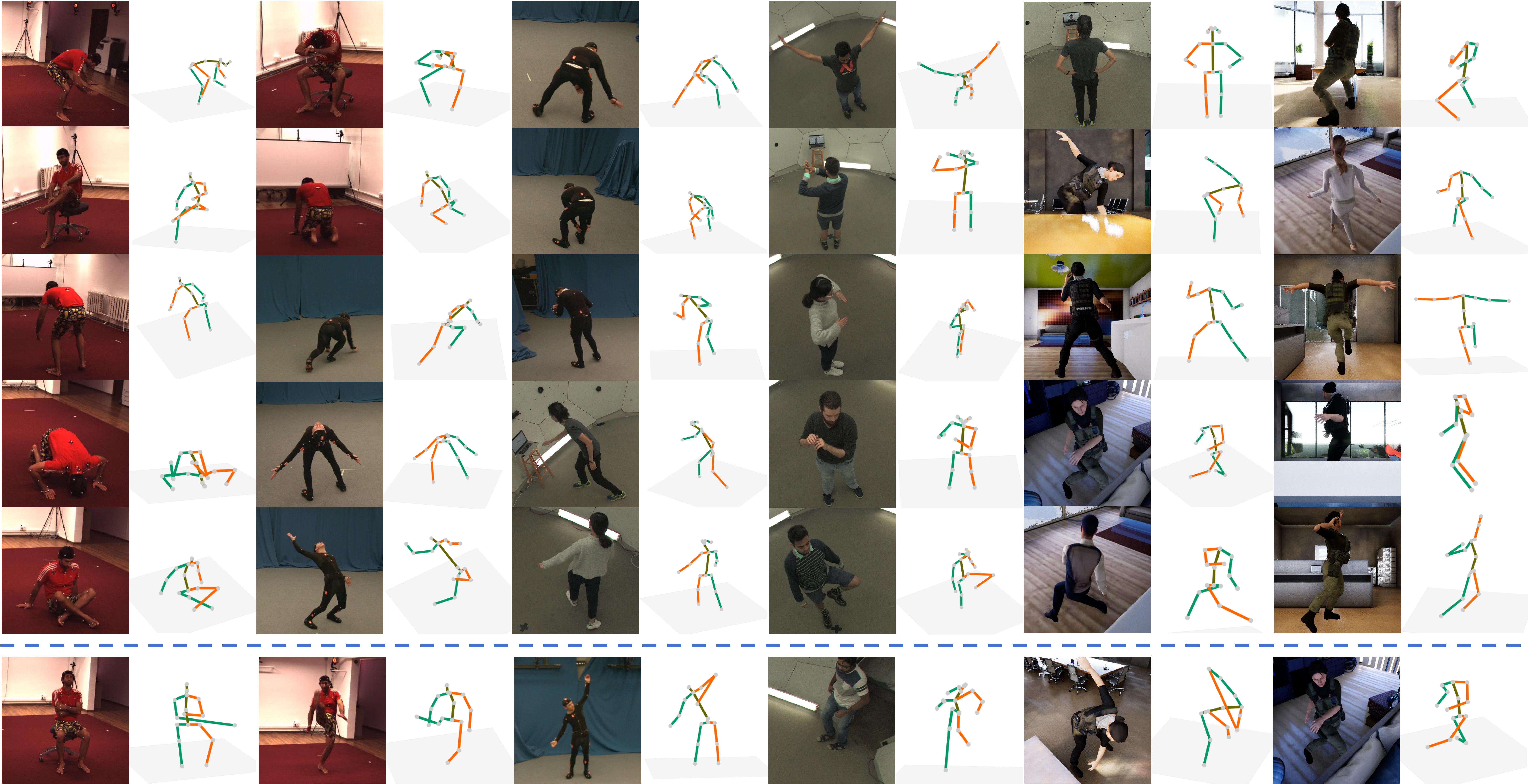}
    \caption{We demonstrate some $3$D pose estimation examples obtained by \emph{AdaFuse}. The last row shows some failure cases. }
    \label{fig:qualitative}
\end{figure*}

\begin{table}[]
\centering
\caption{The 3D pose estimation error ($mm$) of the baseline methods and our approach on the Occlusion-Person dataset. We group the the 3D joints by number of occluded views (8 views in all). We show each group's joint number percentage in the second row.}
\label{table:unreal_occluded_mpjpe}
\begin{tabular}{l|ccccc}
\toprule
Occluded Views   & 4    & 3    & 2    & 1    & 0    \\ 
Percentage       &2\%	 & 15\%	& 38\%&	35\%&	10\% \\ \hline
NoFuse           & 82.6 & 70.2 & 59.7 & 33.7 & 13.0 \\
HeuristicFuse    & 30.5 & 19.9 & 15.9 & 13.5 & 11.1 \\
ScoreFuse        & 25.0 & 18.1 & 15.2 & 13.4 & 12.6 \\
RANSAC           & 36.5 & 24.5 & 19.4 & 14.3 & 11.7 \\
AdaFuse  (Ours)  & \textbf{21.7} & \textbf{14.8} & \textbf{12.5} & \textbf{11.5} & \textbf{10.8} \\
\toprule
\end{tabular}
\end{table}

\begin{table*}[]
\centering
\caption{ The $3$D pose estimation errors MPJPE ($mm$) when \emph{AdaFuse} weight prediction network is trained on \emph{Occlusion-Person} or directly trained on the {Evaluation} dataset, respectively. The $2$D pose estimators for generating the initial heatmaps are trained on each Evaluation dataset separately.}
\label{table:unreal_generalize_to_other}
\begin{tabular}{l|cc|cccc}
\toprule
Evaluation   Dataset & \multicolumn{2}{c|}{AdaFuse}                                 & NoFuse                   & HeuristicFuse            & ScoreFuse                & RANSAC                   \\
               & \multicolumn{2}{c|}{Trained on}                              &                          &                          &                          &                          \\
               & \multicolumn{1}{l}{Evaluation Dataset} & Occlusion-Person         &                          &                          &                          &                          \\ \hline
Human3.6M                          & 19.5                     & 19.4                                 & 22.9                    & 21.0                                               & 20.1                                           & 21.8                                        \\
Panoptic 4 views                   & 14.7                     & 14.6                                 & 33.2                    & 22.5                                               & 21.9                                           & 16.9                                        \\
Panoptic 6 views                   & 13.6                     & 13.9                                 & 29.6                    & 19.8                                               & 19.4                                           & 15.5                                        \\
Total Capture                      & 19.2                     & 20.1                                 & 29.4                    & 20.0                                               & 20.5                                           & 20.5                                       \\
\toprule
\end{tabular}
\end{table*}

\paragraph{Generalization Power}
The only learnable parameters in our fusion approach are in the appearance embedding and geometry embedding networks. In this section, we evaluate whether the {\emph{AdaFuse} weight prediction network} learned on \emph{Occlusion-Person} can be directly applied to the other datasets. {In particular, we append the \emph{AdaFuse} weight prediction network learned on \emph{Occlusion-Person} to the $2$D pose estimators trained on each dataset itself as the final model for evaluation}. Table \ref{table:unreal_generalize_to_other} shows the $3$D pose estimation results on various datasets. We find that the fusion network learned on the synthetic \emph{Occlusion-Person} dataset achieves 
{similar} performance on the three realistic datasets compared to the networks learned on each of the target dataset, respectively. The promising results validate that the fusion model has strong generalization power. It is also worth noting that our approach can naturally handle different numbers of cameras for two reasons. First, the parameters in the appearance embedding network and the geometry embedding network are shared for all camera views. Second, the ``Mean'' operator in the geometry embedding network makes it independent of the number of views as shown in Figure \ref{fig:appearance} and Figure \ref{fig:geometric}. In summary, \emph{AdaFuse} is ready to be deployed in new environments of different camera poses without additional adaptation.

\subsection{Results on Total Capture}
We report the $3$D pose estimation results on the Total Capture dataset in Table \ref{table:mpjpe_totalcapture}. It is worth noting that some methods also use IMUs in addition to the multiview images. We can see that our approach outperforms all of the previous methods. We notice that the error of our approach is slightly larger than LSTM-AE \citep{trumble2018deep} for the ``W2 (walking)'' action of S4,5. We tend to think it is because LSTM can get significant benefits when it is applied to periodic actions such as ``walking''. This is also observed independently in another work \citep{gilbert2019fusing}.

We show some $3$D pose estimation examples in Figure \ref{fig:qualitative}. In most cases, our approach can accurately estimate the $3$D poses. One typical situation where the approach fails is when $2$D pose estimation results are inaccurate for many camera views. For example in the Panoptic dataset, when human begin to enter the dome, they may be occluded in multiple views. In this case, the heatmaps in each view are of low-quality. Therefore the fused heatmaps will also have degraded quality, leading to inaccurate $2$D pose estimations.

\begin{table*}[]
\center
\caption{The $3$D pose estimation errors MPJPE ($mm$) of different methods on the Total Capture dataset.}
\label{table:mpjpe_totalcapture}
\begin{tabular}{l|  c | c | c c c c c c  c}
\toprule
Methods  & IMUs & Temporal & \multicolumn{3}{c}{Subjects(S1,2,3)} & \multicolumn{3}{c}{Subjects(S4,5)} &  \multicolumn{1}{c}{Mean} \\ 
& & & W2 & A3  & FS3 & W2 & A3 & FS3 & \\ \hline

\citep{trumble2017total} & \checkmark & \checkmark &  48.3 & 94.3 & 122.3 & 84.3 & 154.5 & 168.5 & 107.3 \\
\citep{wei2016convolutional} & & & 79.0 &  106.5&  112.1& 79.0 & 73.7 & 149.3 & 99.8\\
\citep{gilbert2019fusing} & \checkmark & &  19.2 & 42.3 & 48.8 & 24.7 & 58.8 &  61.8 & 42.6\\
\citep{trumble2018deep} & & \checkmark & {13.0} & 23.0 & 47.0 & \textbf{21.8} & 40.9 & 68.5 & 34.1 \\
\citep{qiu2019cross} & & & 19 & 21 & 28 & 32 & 33 & 54 & 29 \\
 \hline

NoFuse    &  & & 15.9  & 18.5 & 29.9 & 33.9  & 33.8  & 60.0  & 29.4  \\
HeuristicFuse &  & & 7.8   & 11.6 & 19.6 & 23.3  & 26.9  & 44.8  & 20.0  \\
ScoreFuse  &  & & 9.7   & 13.1 & 19.9 & 23.9  & 27.2  & 41.4  & 20.5  \\
RANSAC  &  & & 8.4   & 11.6 & 20.5 & 23.3  & 27.2  & 45.7  & 20.5  \\
AdaFuse (Ours)  &  & & \textbf{7.2} & \textbf{10.8} &\textbf{18.5} &{22.8} & \textbf{26.6} &\textbf{42.9} &\textbf{19.2} \\

\toprule
\end{tabular}
\end{table*}

\section{Summary and Future Work}
\label{sec:conclusion}
We present a multiview fusion approach \emph{AdaFuse} to handle the occlusion problem in human pose estimation. \emph{AdaFuse} has practical values in that it is very simple and can be flexibly applied to new environments without additional adaptation. In addition, it can be combined with any $2$D pose estimation networks. We extensively evaluate the effectiveness of the approach on three benchmark datasets. The approach outperforms the state-of-the-arts remarkably. We also construct a large scale human dataset which has severe occlusion to promote more research along this direction. Our next step of work is to leverage temporal information to further improve the pose estimation accuracy.

\bibliographystyle{spbasic}      
\bibliography{ref.bib}   
\end{document}